\pdfoutput=1
\documentclass[12pt]{article}

\usepackage[margin=1in]{geometry}
\usepackage{setspace}
\usepackage{times}   
\usepackage{graphicx}
\usepackage{amsmath}
\usepackage{footmisc}
\usepackage{floatrow}
\usepackage{hyperref}
\usepackage{pdfpages}
\hypersetup{pdftitle={Double-Edged Sword of Mediated Visibility: How Visual Framing Undermines Congresswomen's Perceived Competence},
            pdfauthor={Bryce J. Dietrich, Hyein Ko, Myriam Shiran}}
\usepackage[natbibapa, nodoi]{apacite} 

\author{%
  Bryce J. Dietrich\\
  {\small Purdue University}\\
  \and
  Hyein Ko\\
  {\small Case Western Reserve University}
  \and
  Myriam Shiran\\
  {\small The Ohio State University}%
}
\title{Double-Edged Sword of Mediated Visibility: How Visual Framing Undermines Congresswomen's Perceived Competence}

\date{}

\begin{document}

\maketitle

\begin{abstract}
\doublespacing

Women's presence in Congress has stalled below 30\%. That institutional underrepresentation is mirrored by their limited visibility on televised news, where visual presentation can shape perceived competence and authority. Yet research tells us more about whether congresswomen appear than how they are visually framed. Facial recognition analysis of 695,464 image-text segments from CNN and Fox News (2011--2021) reveals that congresswomen appear disproportionately in split-screen rather than solo shots. Two pre-registered experiments with 6,220 participants show that, in static frames, split-screen framing reduces congresswomen's perceived political competence, but not congressmen's. In dynamic videos, the competence penalty disappears; instead, outraged language reduces a congresswoman's perceived warmth roughly three times as much as a congressman's. Suggestive evidence ($p = .057$) indicates that women viewers report greater external political efficacy after watching a congresswoman appear alone. We conclude that visual framing shapes both congresswomen's mediated visibility and their perceived capabilities.

\end{abstract}

\pagebreak
\section{Introduction}

After a decade of marginal gains---from roughly 18\% in 2013 to 28\% in 2023---women's congressional representation plateaued in 2024 and remains below 30\% as of 2026 \citep{conroymarriner2024}. This stagnation unfolds against a backdrop of research documenting congresswomen's underrepresentation on television news \citep[e.g.,][]{andrichdomahidi2024,VanDerPasAaldering2020}, a key arena where political actors shape public perceptions about their competence and authority through mediated visibility \citep{Thompson2005}. That research, however, largely measures visibility through text (who is named, quoted, or mentioned), so it cannot tell us who actually appears on screen or in what visual form. 

Recent evidence suggests that how politicians are visually framed on screen may reinforce or undermine these perceptions as powerfully as the message they convey \citep{kleincoleman2022}. This presents a paradox: congresswomen need mediated visibility to establish their political competence, yet the way they are shown may work against them. Despite this, we know little about how congresswomen are actually presented on screen. We address this gap using facial recognition analysis of CNN and Fox News broadcasts (all programming the TV News Archive captured from the two networks between 2011 and 2021, rather than a selected set of shows) and two pre-registered experiments.

Women's presence on Capitol Hill and the airwaves is fundamental to symbolic representation, defined as how women's presence in positions of power shapes both broader attitudes toward their political competence and women's own sense of external political efficacy \citep{Lawless2004,ReingoldHarrell2010}. These effects, however, depend on more than women's numeric presence in politics; they also require visibility and public recognition of women's political inclusion, which take place primarily through mediated exposure \citep{CampbellWolbrecht2006}.

Yet visibility itself can be a ``double-edged sword'' \citep[p.335]{brighenti2007}. While television appearances offer congresswomen a platform to contest and construct social meanings about their political competence \citep{Thompson2005,Mansbridge1999}, elements largely out of their control—such as how they are visually framed—can undermine its intended impact. 

How congresswomen are made visible on television therefore matters for the symbolic effects of their political representation. We ask three related questions: (1) Do congresswomen receive the same access to mediated visibility on television as congressmen? (2) When they appear, how are they visually framed? (3) How do these framing techniques influence viewers' perceptions of congresswomen—and women viewers' perceptions of themselves? We focus particularly on split-screen shots, which create visible divisions in an otherwise unified frame; they are among the most prevalent visual framing techniques on cable news and heighten the conflict viewers perceive \citep[e.g.,][]{cho2009split}, whereas solo shots allow a political figure to command full viewer attention and project authority \citep{stewart2021visual}. Yet we are aware of no study that has explored split-screen shots on cable news broadcasts.

Ultimately, we find that (1) congresswomen appear on cable news more often than their share of seats would predict, but (2) these appearances are disproportionately staged as split-screens, typically pairing the congresswoman with a male colleague. Two pre-registered experiments then trace the consequences. When viewed as a static frame, (3) split-screen presentation lowers perceptions of a congresswoman's competence, with no similar effect for congressmen. (4) In a video format, we no longer detect this penalty, suggesting that what a legislator says can override how she is framed. The verbal content, however, carries its own cost: (5) the same remarks delivered in an outraged rather than a measured tone reduce a congresswoman's perceived warmth substantially more than a congressman's. Finally, in the dynamic setting, we find suggestive evidence that women viewers who watched a congresswoman appear alone reported greater external efficacy, although the estimate is imprecise ($p = .057$); an exploratory analysis suggests this movement comes from solo appearances delivered in a measured register.

We argue the third and fourth findings reflect two sides of the same interpretive process: a static split-screen leaves open whether the congresswoman is being challenged, and viewers resolve that ambiguity through gendered expectations about conflict, whereas a live exchange shows viewers whether she is. Indeed, viewers who watched the dynamic exchange perceived no more contestation in the split-screen than in the solo shot. The movement in efficacy, meanwhile, is about whether the women watching a congresswoman come to feel more capable themselves.

Taken together, these findings suggest the disadvantage congresswomen face on television has less to do with how often they appear than with how they are shown when they do. If women's presence in positions of power is meant to signal their capability for leadership, then a convention that routinely stages congresswomen opposite a man, and that costs them competence whenever the appearance is reduced to a still, works against the symbolic gains their visibility is supposed to produce. The one condition in which we observe any movement in women viewers' efficacy—the solo frame—is also the one in which congresswomen remain furthest from parity with their male colleagues.

\section{Literature Review}
\label{sec:review}
\subsection{Mediated Visibility, Television, \& Symbolic Representation}

For women historically excluded from politics, greater political inclusion creates new social meanings about who belongs in politics and who is capable of governing, challenging gender norms long used to justify women's exclusion \citep{Mansbridge1999,Sapiro1981}. Seeing more women in positions of power can foster a sense of political empowerment and efficacy by demonstrating that gender need not bar them from political leadership (\citealp[p.651]{Mansbridge1999}; \citealp{Stauffer2021}).

However, empirical evidence on these symbolic effects in the United States remains inconsistent. Some studies find that a higher presence of women in Congress improves political engagement, participation, and efficacy among women \citep[e.g.,][]{Schwindt-BayerMishler2005}, while others report no such effects \citep[e.g.,][]{Lawless2004}; evidence for gender-affinity effects, in which women citizens respond most directly to women candidates and officeholders, is similarly mixed \citep{Dolan2006,KraftDolan2023}. One explanation holds that symbolic effects do not follow automatically from descriptive representation: citizens must see and register changes in women's representation, primarily through mediated visibility \citep{ReingoldHarrell2010}.

`Mediated visibility'---the process by which political actors gain recognition through their media appearances---is therefore a key mechanism by which symbolic effects materialize. Political actors do not come to `stand for' a group simply by holding office; they actively construct symbolic meanings through their actions and appearances \citep{Mansbridge1999}. Symbolic effects are accordingly stronger and more consistent when citizens know women representatives through media exposure \citep{LadamHardenWindett2018}.

These symbolic effects operate on two partially independent dimensions. The first concerns how viewers evaluate the politician herself—whether seeing a congresswoman in a position of authority leads viewers to judge her, and women in politics more generally, as capable. The second concerns viewers' attitudes toward the political system—whether a congresswoman's presence makes women watching believe that government can be responsive to people like them \citep{Stauffer2021}. There is little reason to expect a single appearance to move both dimensions in tandem, because each judgment draws on different aspects of the on-screen depiction. Judgments of competence depend on performance: when an appearance supplies evidence of how a politician carries herself, viewers draw on it, but when that evidence is minimal—a still image, a passing glimpse, or a muted clip—they fall back on compositional cues as heuristics for credibility and expertise, especially when time and information are limited \citep{grabe2009image}. Orientations toward the political system rest instead on recognition, on whether a congresswoman registers as a person speaking to the viewer rather than as a figure arranged within the frame \citep{ReingoldHarrell2010}. This distinction between being seen as capable and feeling capable oneself implies that compositional cues should weigh most heavily where an appearance provides little else. It also leaves open whether effects observed in thin, static slices of a broadcast carry to the richer viewing experience of the broadcast itself.

However, mediated visibility is not inherently empowering. As high-visibility spaces, mass media shape public perceptions of political competence and power structures not simply through who appears, but through how they are visually presented \citep{brighenti2007}. Both carry consequences: exclusion alone condemns political actors to obscurity and even ``a kind of death by neglect'' \citep[p.49]{Thompson2005}. Yet existing work on congresswomen speaks almost entirely to the first. For women members of Congress (MCs), the struggle for visibility is particularly acute on television, which remains among the most widely used sources of political news for American adults \citep{pew2024electionnews}. Yet most studies of congressional news coverage focus on print media \citep[e.g.,][]{lucas2017gender}. Within the smaller body of work on television, women appear less frequently and receive less airtime on Sunday talk shows \citep{baitinger2015meet}, and a meta-analysis confirms that television remains a comparatively disadvantaged platform for women \citep{VanDerPasAaldering2020}. These estimates rest on guest lists, transcripts, and program-level counts, and the U.S. evidence draws heavily on Sunday morning programming—designs that can establish whether a congresswoman was on the air, but not what viewers actually saw when she was. Where visual depictions have been examined directly, gendered patterns emerge: analyzing candidate imagery across 28 countries, \citet{jungblutHaim2021} document systematic visual gender stereotyping in how women and men politicians are shown, though in still news photographs rather than broadcast appearances.

Like textual framing, visual framing highlights and organizes aspects of reality to promote certain meanings \citep{entman1993}, but images are processed more quickly and evoke stronger emotions \citep{brantnerlobingerwetzstein2011}. When text and image conflict, visual information typically dominates interpretation—in part because audiences are often unaware of these effects and rarely scrutinize images as critically as verbal arguments \citep{gibsonzillmann2000}. Studies show that visual frames affect opinion formation, comprehension, and evaluations of political figures \citep{fahmykim2008}, and that the visual presentation of women candidates in particular shapes support for them \citep{bauerCarpinella2018,carpinellaBauer2019}. On cable news, the most consequential of these choices may also be among the simplest: whether a congresswoman is shown alone or alongside someone else.

Visibility is not neutral; it is structured by who controls the means of exposure and by conventions that render certain bodies legible as authority figures \citep{banet2018empowered}. We conceptualize mediated visibility as occurring in three interdependent stages, each influenced by gender: exposure and framing concern whether a congresswoman appears and how she is shown, while interpretation concerns how audiences judge what they see. Differentiating the third stage matters because what viewers make of a shot depends on what other information the appearance provides. In this account, whether symbolic representation succeeds depends on the second and third stages.

\subsection{Split-Screen as a Style of Visibility \& Its Consequences}

Among visual framing techniques, split-screen presentations warrant particular attention given their prevalence in cable news coverage. By visually juxtaposing politicians, split-screen shots can create a sense of opposition or heighten the level of conflict discerned by viewers \citep{cho2009split,scheufele2007my,wicks2007does}, a tendency compounded on programs that favor highly partisan members as guests \citep{baitinger2015meet}. The evidence comes largely from presidential debates. \citet{stewart2021visual} argues that occupying adjacent visual space prompts comparison and invites competitive evaluation, whether candidates stand side by side on a stage or are joined within a split-screen. Using a $2 \times 2$ design, \citet{cho2009split} find that viewers perceived more incivility in the split-screen than in the single-screen condition, with \citet{scheufele2007my} and \citet{wicks2007does} reporting converging results. In this literature, the frame's consequences are understood to run through perception; viewers see a divided screen and infer a contest. Whether a given presentation actually produces that perception is an empirical question, and one we measure directly.

Solo shots work in the opposite direction. When a candidate is shown alone, especially in head-and-shoulders framing, that individual dominates viewers' perceptions \citep{mutz2015your} and can be ``cast'' in an artificially intimate manner \citep{sullivan1988happy}, which may shape which figure viewers regard as the more appropriate leader \citep[p.549]{stewart2017visual}. The gendered stakes are visible in a study of the third Trump--Clinton debate: among 139 undergraduates who watched in either split-screen or switched-feed format (the latter comprising predominantly solo shots), Hillary Clinton was rated significantly higher on \emph{strong}, \emph{competent}, and \emph{intelligent} when shown alone \citep{stewart2017visual}. 

Cable news, however, is not a debate. Networks have less reason to stay neutral and more to highlight conflict, and a politician's \textit{on-screen opponent} is often her own host rather than an avowed adversary. Thus, the same split-screen may contain a hostile cross-examination or a cordial interview. In a debate, watching largely confirms what the divided frame implies---the speakers are in fact opponents---whereas on cable news it may disconfirm it. Dynamic exposure does not make viewers immune to compositional cues, but it supplies evidence against which those cues can be checked; a static frame supplies none.

For congresswomen, the ambiguity of the divided frame is not resolved neutrally. A split-screen invites viewers to interpret a congresswoman's participation as contested rather than routine, and the consequence falls on competence, because a frame that marks a speaker's position as contested presents her authority as provisional. How often that inference goes unchecked is an empirical matter, and cable news is consumed in strikingly heterogeneous ways: some viewers watch attentively from start to finish, while many more monitor the news in passing, catching brief fragments of coverage rather than following a conversation whole \citep{costerameijer2015checking}. An increasing number encounter broadcast content not through full programs but through social platforms where news clips are short, often muted \citep{newman2025dnr,kalogeropoulos2018newsvideo}, and where screenshots and thumbnails circulate separately from the actual exchanges \citep{matatov2022stop,brennen2021beyond}. Nor are these circulating fragments neutral previews; thumbnails often form the first impression that decides whether full content is watched at all \citep{poudelCakmak2026}, and the imagery selected carries gendered stereotypes of its own \citep{chenDuanKim2024}. Attentive viewers can tell whether an exchange is adversarial; casual viewers see only what the frame shows and must infer the rest. When an encounter carries little of the underlying exchange, the divided frame becomes the viewer's only account of what took place.

Composition is not the only gendered mechanism at work in a television appearance, however. An appearance can also trigger an agentic penalty \citep{schneideretal2022}, whereby women are sanctioned for displaying the very assertiveness that political leadership is thought to require. The stereotype content model holds that social perception is organized along two fundamental axes—warmth and competence—and that prescriptive stereotypes assign women the former while reserving agency for men \citep{fiske2002model,rohrbach2025}. When women display the assertiveness, anger, or dominance expected of political leaders, the resulting sanction therefore affects not only judgments of their competence but also their perceived warmth, with women who express the same anger as men seen as colder and less deserving of support \citep{eagly2002role,okimoto2010price,brescoll2016leading}. Unlike the compositional cues described above, this penalty attaches to what a congresswoman says and how she says it, and so follows her across formats.

Gendered evaluation of this kind can operate in two analytically distinct ways, which existing work on visual framing rarely distinguishes. The first is a double standard, in which an identical presentation carries different consequences depending on whether the person shown is a woman or a man \citep{hargrave2023double}. The second is a double bind, in which the agentic behavior expected of leaders is itself sanctioned when women display it, so that the same words carry different costs depending on who speaks them \citep{okimoto2010price,brescoll2016leading}. The two are easily conflated because both produce gendered penalties, yet they involve different elements of a television appearance: the double standard turns on the visual form of the appearance, the double bind on the register of what is said.

\section{Theoretical Expectations}

How a congresswoman is made visible---her style of visibility---can constrain the social meanings she seeks to construct and convey. One widely used visual framing technique on cable news programs is the split-screen shot, which visually juxtaposes politicians in ways that the debate literature finds heighten perceptions of conflict and confrontation. This framing technique may be particularly problematic for congresswomen: when a divided frame implies contest, viewers lean on gendered expectations to decide what they are seeing, and the congresswoman comes across as a leader whose authority is in question. Conversely, solo shots let politicians hold viewers' full attention and project authority without triggering competitive evaluations. This discussion leads to our first hypothesis:

\begin{quote}
    \emph{H1: Congresswomen who appear in solo shots will be perceived as more competent than those who appear in split-screen presentations.}
\end{quote}

How far this first expectation reaches, however, depends on the informational setting in which the judgment is made. If viewers judge competence from what they can actually watch a politician do, as we argued above, then the compositional cues of the frame should have their greatest effect when a broadcast appearance is encountered as a static excerpt, and their smallest when the appearance unfolds in time and the performance can be observed. Whether the split-screen penalty carries from static frames to dynamic viewing is therefore an open question for our account, and one our experiments answer directly.

The second dimension distinguished above supplies an expectation of its own. Beyond perceptions about congresswomen's competence, seeing a congresswoman in a position of authority may also strengthen women viewers' external political efficacy---their belief that political institutions can be responsive to their concerns. However, this symbolic benefit may depend on how she is visually presented, leading to our second hypothesis:

\begin{quote}
    \emph{H2: Women viewers who see congresswomen in solo shots will report higher external political efficacy compared to women viewers who see congresswomen in split-screen presentations.}
\end{quote}

The same informational logic that bounds our first expectation applies here as well. If external efficacy rests on the kind of recognition we described above, a person speaking rather than a figure arranged within a shot, then the solo shot's benefit should emerge most clearly where the appearance unfolds with voice and motion, and least where it is reduced to a static excerpt.

Thus far, these expectations treat the message congresswomen convey as fixed. Yet the literature reviewed above points to the emotional tone of their words as a second, analytically separable source of gendered evaluation. It also suggests two distinct processes through which that register may interact with visual framing. If split-screen presentations harm congresswomen by coding their claims as contested, then outraged language should supply confirmatory evidence of that contest: the same forceful words that count as advocacy in a solo shot should count as evidence of conflict in a split-screen, concentrating the penalty where visual contest and outraged words coincide. We refer to this process as `contested advocacy,' and it leads to our third hypothesis:

\begin{quote}
    \emph{H3: Congresswomen who express outrage will incur a larger split-screen competence penalty than congresswomen who speak in a measured register.}
\end{quote}

The agentic-penalty literature suggests a rival process. If the sanction attaches to the display itself, not to its visual context, then outraged words should be costly for a congresswoman wherever she appears, because those words violate prescriptive expectations of communality, whatever the screen shows \citep{okimoto2010price,brescoll2016leading}. Under this account the penalty should follow the gender of the speaker regardless of the arrangement of the frame, and it should register on perceived warmth at least as strongly as on perceived competence \citep{fiske2002model}. We refer to this process as `agentic backlash,' and it leads to our fourth hypothesis:

\begin{quote}
\emph{H4: Congresswomen who express outrage will suffer a greater loss of perceived warmth than congressmen who express identical outrage.}
\end{quote}

These final two expectations cannot both be right, and they disagree most sharply about a congresswoman who expresses outrage while appearing alone: if forceful advocacy is penalized only when the screen codes it as contested, she should escape sanction, whereas if the backlash attaches to the words themselves, she should not. Each account also identifies a distinct first link in its causal chain (perceived contestation for contested advocacy, perceived warmth for agentic backlash), so measuring both perceptions alongside our outcomes allows the data to indicate which process penalizes congresswomen. Because our experiments cross the gender of the member of Congress with both the visual format and the emotional register of otherwise identical appearances, they also separate the double standard from the double bind described above.

\section{Data and Methods}
\subsection{TV News Archive Data}

Since 2007, the Internet Archive has recorded over 1,417,000 national and local television news programs in a database called the \emph{TV News Archive}. For the purposes of this study, we subset the data using the \cite{PearsonDancey2011a} dictionary, which includes the following terms: ``woman,'' ``women,'' ``woman's,'' ``women's,'' ``girl,'' ``girl's,'' ``girls,'' ``girls','' ``female,'' ``female's,'' ``females,'' ``females','' ``servicewoman,'' ``servicewoman's,'' ``servicewomen,'' and ``servicewomen's.'' We focus on CNN and Fox News because they are two of the most-watched cable news networks with sustained TV News Archive coverage, and because they sit on opposite partisan sides, a pattern common to both is unlikely to reflect the politics of either network alone. 

The study window, from May 13, 2011, to March 16, 2021, spans the 112th through 117th Congresses, and it holds substantive interest for symbolic representation: it opens with women holding roughly 17 percent of congressional seats, captures the record influx of women elected in 2018, and closes just after the seating of the 117th Congress, then the high-water mark of the trajectory described in our introduction \citep{conroymarriner2024}. Six Congresses also give the observational analysis breadth across political contexts: the timeframe covers Democratic and Republican control of each chamber and spans parts of three presidencies, so that no single election cycle, majority, or administration drives the patterns we document. Finally, these six Congresses comprise a total of 953 members whose images our facial recognition model was trained to identify, with the exact endpoints set by the availability of paired image-caption data at the time of collection.

Once these terms were applied to CNN and Fox News broadcasts during this period, we obtained 374,643 and 320,821 image-text segments, respectively. Each segment corresponds to one minute of a cable news broadcast, and the Appendix reports the number of segments returned by each term for each network.

\subsection{Small-Sample Facial Recognition Using Siamese Neural Networks}

We use facial recognition to identify television appearances of MCs. Previous studies often rely on either automatic or manual coding of mentions in news broadcasts to measure visibility \citep{WagnerGruszczynski2018,baitinger2015meet}. However, these mentions do not capture whether a politician was physically shown, interviewed, or merely referenced in commentary. This gap is especially problematic in assessing the visibility of women MCs, whose marginalization often manifests visually as well as verbally. In addition, text-based approaches struggle with name ambiguity: for MCs with common names, such as Mike Rogers (R-AL), a mention need not refer to the politician at all, leading to speaker misidentification. Facial recognition avoids this limitation and allows us to explore new questions centered on visual framing.

We developed a custom Siamese Neural Network that identifies who is on screen from small training samples. This approach is widely used in computer science when training data are limited (for review, see \citeauthor{Chicco2021}, \citeyear{Chicco2021}), but has yet to be introduced to political science. As the name implies, the algorithm relies on two identical subnetworks that share architecture, parameters, and weights and learn from image pairs---positive pairs share the same label, negative pairs do not---rather than from individually annotated images. Pairwise training dramatically reduces the number of new training examples needed: conventional convolutional neural networks require multiple images per category, an impractical demand when identifying MCs, given the limited availability of images for rank-and-file legislators and the large number of categories (e.g., 535 per Congress), which hinders class balance and degrades model performance \citep{JapkowiczStephen2002}. The Appendix presents the architecture and a worked example of the output.

We trained the algorithm for 500 epochs on a high-performance computing cluster (see Appendix for details), which took approximately 8 hours. To illustrate, the Appendix walks through a comparison of Nancy Pelosi with the actress Emma Watson: Pelosi compared with herself yields a predicted contrastive loss near zero (0.00032), while the same image compared with Watson's yields a high value (1.71736), correctly separating two faces that share head position and facial structure.

With the model trained, we identified whether an extracted face belonged to a member of Congress. Since the algorithm estimates facial similarity, extracted faces that score higher against our legislator dataset are more likely to be members of Congress. To calibrate, we compared each face in our data with 20 randomly selected celebrity images, matched to the gender and facial positioning of a given MC. Using the mean and standard deviation of these comparisons, we established 95-percent confidence intervals; a face exceeding this interval is classified as a member of Congress. This process was repeated for all 953 legislators from the 112th to 117th Congresses.\footnote{In Section 2 of the Appendix, we provide a step-by-step example of this calculation.}

We validate our facial recognition algorithm in two ways (see Appendix Section 3 for more information). First, we replicated the calibration comparison for every member of Congress: predictions comparing an MC's face to itself did not significantly differ from zero ($\bar{x} = 0.0003162278$), while comparisons against the faces of 20 randomly selected celebrities who share the MC's gender and facial positioning were significantly higher ($\bar{x} = 2.12483$, $t = 245.48$, $df = 19074$, $p < 0.0001$), indicating the model reliably distinguishes legislators from matched non-legislators.

Second, we used a collection of more than 120,000 LexisNexis transcripts from the same period (2011--2021) to establish convergent validity with a text-based approach \citep{chin2024convergent}. Focusing on transcripts where MCs make an appearance, we matched those segments to one-minute blocks of closed-captioning data from the TV News Archive, using the MiniLM sentence transformer to compute semantic similarity (captions carry typos, omissions, and no speaker identification) and recorded the number and gender of MCs speaking in each segment. As shown in the Appendix, this text-based measure yields the same substantive results, which indicates the patterns we document are not artifacts of our facial recognition pipeline. Still, the two approaches are not interchangeable. A transcript can establish that a congresswoman spoke during a broadcast but not whether she appeared on screen or how the shot was composed, so the text-based measure serves as a check on our approach rather than a substitute for it.

With the legislator faces identified, we then subset our data to instances where either a single legislator appeared (solo shot) or two legislators appeared on screen (potential split-screen shot). This produced 100,816 CNN frames and 125,251 Fox News frames in which one legislator appeared, and 32,642 CNN frames and 34,281 Fox News frames in which two legislators appeared. To determine whether these two-person frames were presented as split-screen shots, we implemented a computer vision pipeline using CLIP, a vision-language model developed by OpenAI that is widely applied to ``zero-shot'' image classification \citep{radford2021learning}. Rather than training a custom classifier, which would require extensive manual coding, we constructed textual descriptions of prototypical ``split-screen'' and ``single-camera'' shots (e.g., ``a TV split-screen of two people'' versus ``a single camera shot of one scene'') and calculated similarity scores between each frame and these prompts, estimating the probability that a frame was a split-screen shot without supervised training. Applying a pre-specified decision threshold of 0.70, we identified 27,658 CNN frames and 25,519 Fox News frames as split-screen shots, which form the basis for our subsequent analyses. Additional details on this pipeline are in the Appendix.

\section{Results}

\subsection{How are Congresswomen Visually Framed?}

We first use the pipeline just described to establish that congresswomen are disproportionately featured in split-screen shots. Across CNN and Fox News, there are 53,177 split-screen frames (CNN = 27,658; Fox News = 25,519), and congresswomen appear in 51.6 percent of them (CNN = 13,146; Fox News = 14,274). By comparison, among 226,067 frames showing a single legislator, women appear in only 32.6 percent (CNN = 30,390; Fox News = 43,366). A chi-square test confirms that the gender imbalance in split-screen coverage is highly significant ($\chi^2 = 6682.6, df = 1, p < 0.0001$).

Not all split-screen configurations are equally common. The least frequent pairing features two congresswomen (CNN = 8.9 percent; Fox News = 12.9 percent), while the most frequent pairing involving a congresswoman places her beside a congressman, accounting for roughly two-fifths of all split-screen frames (CNN = 38.6 percent; Fox News = 43.0 percent). This pattern holds even though women as a group are the topic of discussion in every segment we analyze.

\begin{figure}
     \caption{Women legislators are significantly more likely to appear in split-screens on CNN and Fox News}
     \centering \includegraphics[width=.7\linewidth]{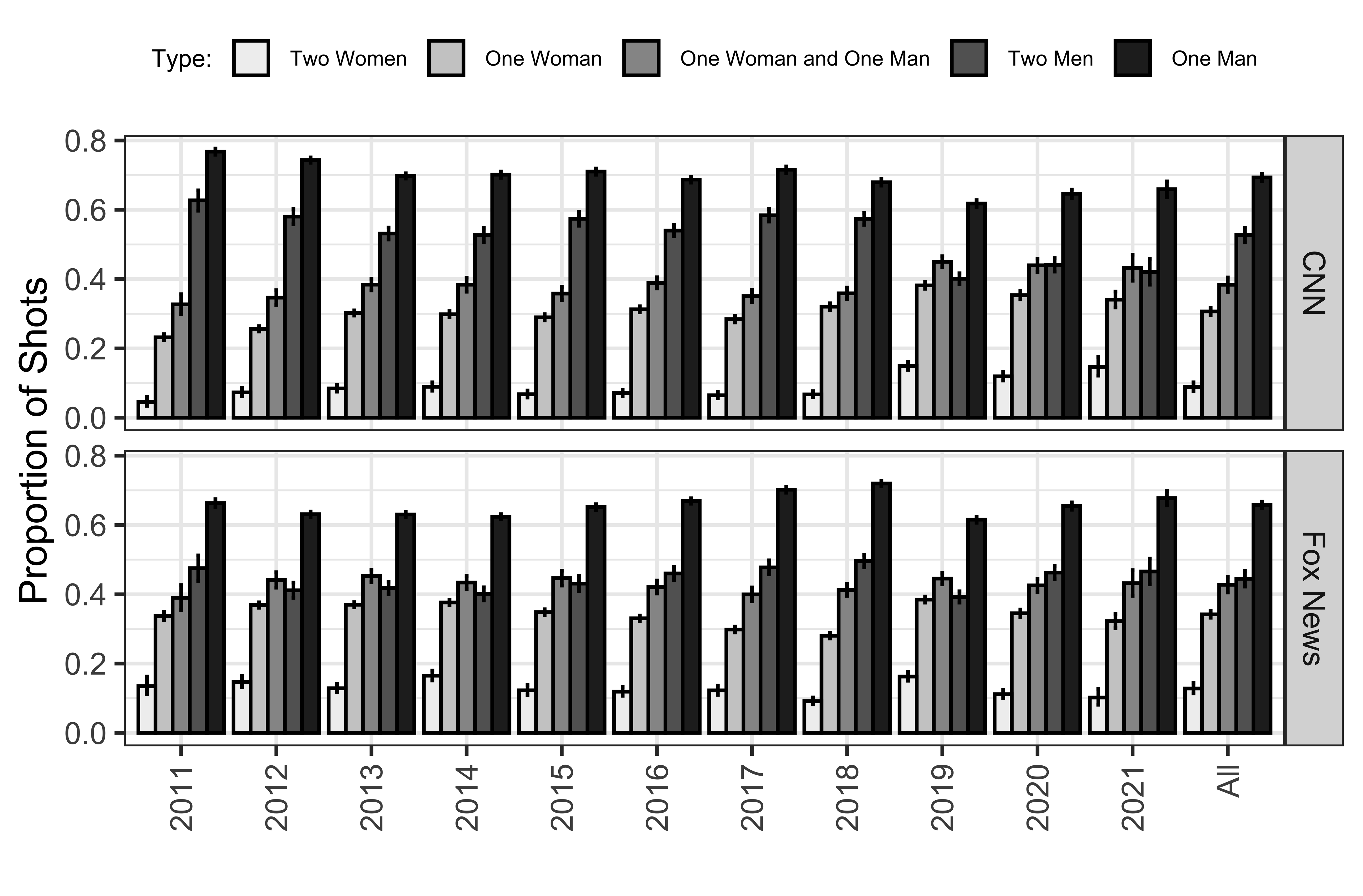}
     \floatfoot{\setstretch{1}\emph{Note}: Colors indicate frame type, ordered from least to most likely. The top panel shows CNN broadcasts, the bottom Fox News. Details on the facial recognition algorithm are in Section 1 of the Appendix.}
     \label{fig:fig5}
 \end{figure}

As shown in Figure \ref{fig:fig5}, these discrepancies remain consistent over time. Across the study period, the most common visual frame overall is a solo appearance by a congressman (69.9 percent of CNN's solo frames; 65.4 percent of Fox News's). From 2011 to 2021, men outnumbered women in Congress by about 3.9 to 1, yet the men-to-women ratio is just 2.1 in solo frames (CNN = 2.3; Fox News = 1.9) and 1.7 in split-screens (CNN = 1.9; Fox News = 1.6). Despite Fox News's partisan lean and the smaller number of Republican women, the disparity is no stronger there than on CNN, which suggests a production convention shared across the partisan spectrum.

\subsection{When Congresswomen Appear in Split-Screen Shots, Are They Perceived as Being More or Less Competent?}

To test our first hypothesis, we conducted a pre-registered online vignette experiment through Prolific.\footnote{The pre-analysis plan, registered with the Open Science Framework on July 23, 2026 (extending our December 2024 pilot registration), and its dated deviations log are available anonymously at \url{https://osf.io/382xu/overview?view_only=ffd2a273a2104c58b7e1c22351280306}.} Because a design that shows only a congresswoman cannot determine whether the penalty is gendered, this experiment fully crosses the gender of the MC with the shot format. Respondents were assigned to one of six conditions in a $2 \times 3$ design, seeing either a woman or a man MC (1) appearing alone, (2) with a same-gender colleague in a split-screen, or (3) with an opposite-gender colleague in a split-screen. The two solo conditions each comprised 25 percent of the sample and the four split-screen conditions 12.5 percent each, an unequal allocation that enhances the gendered comparison's efficiency within a simple $2 \times 2$ design; the target sample size was fixed in advance by a pre-registered power analysis.

Respondents (N = 2,619 analyzed, 49\% women) completed the six-minute survey on July 23, 2026, receiving \$1.50 for their participation; recruitment was balanced by gender, and participants in the December 2024 pilot (reported in the Appendix) were excluded. Of 2,733 completed responses, two consent decliners (who reached the survey's end without being randomized) are removed under the registered plan. Of the remaining 2,731, the primary analyzed sample retains the 2,619 passing a revised attention rule adopted after the soft launch revealed that attentive respondents often responded to the check's factual content rather than following its embedded instruction, a revision recorded in the registration's deviations log before full launch and before any treatment effects were estimated. All confirmatory estimates are additionally reported under the registered strict rule (N = 1,961) and with no attention exclusion (N = 2,731) in the Appendix. After giving informed consent and answering questions about their demographic background, partisanship, and recent voting history, they were randomly assigned to one of the six treatment conditions.

These treatments consist of six static frames in the style of a CNN cable news broadcast, built from the same four individuals—a woman MC, a man MC, a woman colleague, and a man colleague—with identical graphics and accompanying text in every condition, so that only the gender of the MC and the format of the shot vary across cells. The frames are accompanied by a brief explanation of the subject under discussion, which is healthcare; an example frame, the full set as fielded, and the explanation are provided in Section 5 of the Appendix.

After seeing the frame and reading the text, participants assessed the political capability of the MC, whose frame was re-displayed on each evaluation screen with a green box identifying the target so that all evaluations referenced a specific individual without naming them. To measure perceived competence, participants evaluated how well the MC could address specific issues, from ``Poorly'' to ``Very Well.'' Since the headline in the image relates to healthcare, responses to the healthcare item were benchmarked against the mean rating of three other policy areas (education, national security, and the economy).

If our first hypothesis is correct, the congresswoman should be perceived as better able to handle healthcare---the topic of the treatment---when she appears by herself. Because the same comparison is available for the man MC, the design also affords a direct test of whether any such penalty is shared: if the split-screen imposes a generic cost, he should suffer it in equal measure, whereas the double standard described above implies the loss should fall disproportionately on the congresswoman. Our pre-analysis plan registered this difference in penalties as a directional (one-sided) test at the five percent level, with the two-sided value reported alongside; this is the comparison the unequal allocation described above was designed to favor. 

\begin{figure}
     \caption{Congresswomen are significantly more likely to be seen as competent when they appear by themselves, an advantage congressmen do not share}
     \centering
     \includegraphics[width=\linewidth]{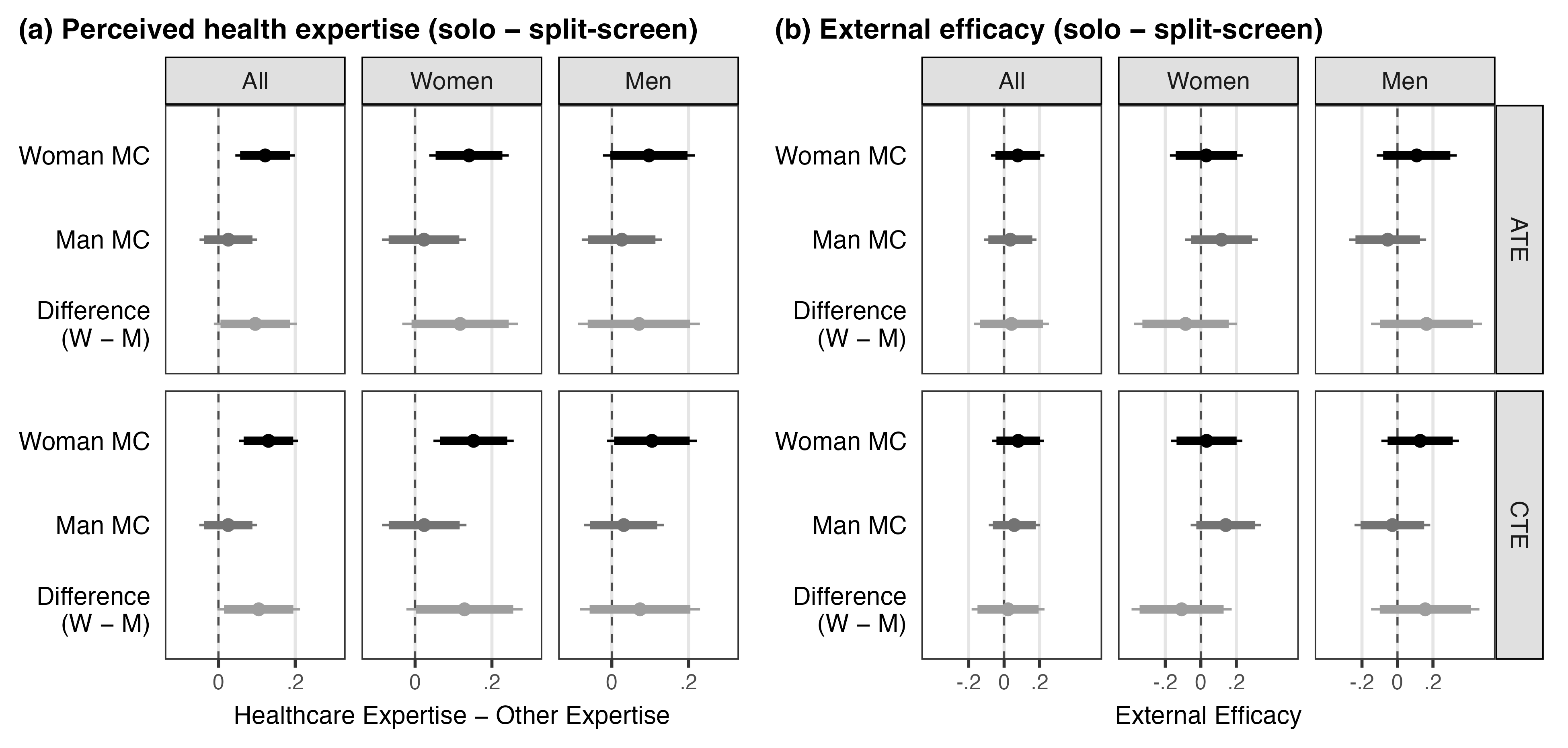}
     \floatfoot{\setstretch{1}\emph{Note}: OLS estimates on the six treatment-cell indicators, with HC2 robust standard errors; contrasts are labeled on each panel's axis, displayed as solo minus split-screen, with the woman MC's contrast set in black. Facet columns are respondent subgroups; ATE panels are unadjusted, and CTE panels add controls for respondent gender, age, education, race, party identification, political interest, and internal efficacy. Panel (a): perceived health expertise (benchmarked against education, national security, and the economy); positive values indicate a split-screen penalty; N$_{all}=2618$, N$_{Women}=1295$, N$_{Men}=1281$ (2601, 1286, 1273 with covariates); tests are two-sided except the difference, registered as one-sided. Panel (b): external efficacy; positive values indicate greater efficacy when the MC appears alone; N$_{all}=2619$, N$_{Women}=1295$, N$_{Men}=1282$ (2602, 1286, 1274 with covariates); all tests are two-sided. Ns are per-outcome complete cases (other-gender respondents in the full sample only). Thicker/thinner lines: 90\%/95\% CIs; full results in the Appendix.}
     \label{fig:fig7}
\end{figure}

In Figure \ref{fig:fig7}, we report both average (ATE) and conditional treatment effects (CTE). In both instances, a single linear regression of the outcome on indicators for the six treatment cells was used, with heteroskedasticity-robust (HC2) standard errors; for both experiments we report coefficients with confidence intervals and exact $p$ values. In the conditional specification, we included the registered pre-treatment covariates: the participant's gender, age, education, race, party identification, political interest, and internal efficacy (see the Appendix for the demographic composition of both samples). Looking to the black lines in the first column of Figure \ref{fig:fig7}a, we find the ATE associated with our treatment (the congresswoman's solo advantage, equivalently her split-screen penalty, pooled over colleague gender) is positive and statistically significant ($\beta = 0.122$, 95\% CI $[0.044, 0.199]$, $p = .002$), implying congresswomen are perceived as more competent when they appear by themselves than in a split-screen. The result holds when the effect of treatment is conditioned on the covariates ($\beta = 0.130$, 95\% CI $[0.053, 0.207]$, $p = .001$). It is larger still when the congresswoman alone is compared to the congresswoman beside a male colleague ($ATE\colon \beta = 0.164$, 95\% CI $[0.071, 0.258]$, $p = .001$; $CTE\colon \beta = 0.181$, 95\% CI $[0.090, 0.272]$, $p < .001$). We also find that the penalty is statistically significant among women respondents ($\beta = 0.140$, 95\% CI $[0.037, 0.244]$, $p = .008$) and positive but not significant among men ($\beta = 0.097$, 95\% CI $[-0.023, 0.216]$, $p = .113$), although these specific comparisons were not pre-registered as hypotheses. The raw responses (five-point item: 3.973 solo, 3.784 split), tabulated per cell in the Appendix for every condition, outcome, and exclusion regime, show the same pattern; every cell mean sits above the midpoint, so the split-screen makes the congresswoman look less competent relative to her solo appearances, not incompetent.

By contrast, the dark gray lines in Figure \ref{fig:fig7}a show that an identically framed congressman suffers essentially no penalty ($\beta = 0.026$, 95\% CI $[-0.049, 0.101]$, $p = .501$). Because no equivalence threshold was pre-registered for this quantity, we report only its confidence interval. The light gray lines report the difference between the two penalties. Consistent with our registered directional expectation, the split-screen penalty is larger for the congresswoman than for the congressman ($\beta = 0.096$, 90\% CI $[0.005, 0.187]$, $p_{\text{1s}} = .041$; two-sided $p = .082$). The estimate is similar under the covariate-adjusted specification ($\beta = 0.105$, 90\% CI $[0.015, 0.195]$, $p_{\text{1s}} = .028$; two-sided $p = .056$) and essentially unchanged across the three registered attention regimes ($p_{\text{1s}} = .041$, $.054$, and $.032$ under the revised, strict, and no-exclusion rules, respectively), suggesting the split-screen penalty is gendered, not a general cost of shared screens.

Collectively, these findings provide strong evidence for our first hypothesis and emphasize the attributional advantages of the solo shot; in the fully crossed design, those advantages accrue only to the congresswoman, marking the penalty as the double standard distinguished in Section \ref{sec:review}. It is also worth noting that women respondents in particular appeared more responsive to solo framing, consistent with social-identity perspectives in which same-gender exemplars reinforce perceptions of capability and belonging within political institutions \citep{eagly2002role}.

Although our results strongly support the first hypothesis, we find little evidence for the second (panel (b) of Figure~\ref{fig:fig7}). External efficacy is the mean of agreement with three items---``Most public officials care what people like me think,'' ``People like me have a say in what the government does,'' and ``No matter whom I vote for, it won't make a difference to what happens in the country'' (reverse-coded)---each scaled from $-3$ to $3$, with higher values indicating greater efficacy. Across all model specifications, the treatment coefficient was never statistically significant at the 0.05 level, though its direction was generally consistent with expectations; the pre-registered equivalence bounds exclude all but small effects. Full diagnostics, as specified in our pre-analysis plan, are in the Appendix.

\subsection[Does the Split-Screen Penalty Persist When the Appearance Unfolds as Dynamic Video?]{Does the Split-Screen Penalty Persist When the Appearance Unfolds as Dynamic Video?}

In every test conducted so far, the broadcast appearance reached viewers as a static image. A static frame cannot show whether the second speaker is interrupting, rebutting, or waiting to speak, so viewers must infer whether the exchange is adversarial. Our second experiment therefore moves the appearance into the dynamic video format and randomly manipulates the emotional tone of the message. 

Register manipulation has an observational warrant. Prior work shows that when the topic is women, congresswomen speak with heightened emotional intensity \citep{dietrich2019pitch}, that women legislators are more likely to use emotive language \citep{gennaro2022emotion}, and that women candidates increasingly employ outrage in their digital appeals \citep{russell2024not}. The coverage in which these appearances are embedded, however, is not itself pitched at that register: scoring the closed-caption text of all 695,464 segments in our corpus against the anger and fear categories of the NRC Emotion Lexicon \citep{mohammad2013emolex}, we find that such vocabulary amounts to only about 29 of every 1,000 words spoken. The words that mark the outraged register itself are rare, suggesting most segments sit closer to our measured script (the Appendix reports this analysis in full and certifies, with the same instrument, that the two scripts introduced below carry the registers they are meant to carry). 

Our treatments therefore test what one is likely to observe on an average cable news broadcast (the measured script) against what one may observe in rare, but perhaps particularly important, instances (the outraged script). An earlier static pilot study, reported in the Appendix, foreshadowed this test but could not run it. Its outraged caption appeared in both of its conditions, holding the register constant, so it could show that the penalty persists but could not identify the effect of outrage itself. Randomizing the register closes that gap.

Specifically, we pre-registered a second experiment, administered through Prolific, employing a $2 \times 2 \times 2$ between-subjects design that crosses the gender of the MC with the format of the shot (solo versus split-screen) and the emotional register of the statement (outraged versus measured).\footnote{The pre-analysis plan, registered under a one-year embargo with the Open Science Framework on July 27, 2026 (again extending our December 2024 pilot registration), and its dated deviations log are available anonymously at \url{https://osf.io/cnk2d/overview?view_only=818e52e8a4be4c34936aed008f28792f}.} Each respondent watched one of the eight resulting segments, each roughly 40 seconds long, presenting an interview in the style of a cable news broadcast. A male anchor (the same in every condition) introduces a fictional MC and closes the segment, a chyron carries the healthcare headline used in our first experiment, and a rolling transcript highlights each phrase as synthesized voices deliver it. A green box marks the MC, as in our first experiment; the constructed faces from that experiment again portray the two MCs. In the solo conditions the screen cuts to whoever is speaking, so the MC delivers the statement alone on screen. In the split-screen conditions the MC and the anchor share a two-shot throughout, the active speaker in color and the listener in grayscale, so the same statement is presented as one side of a visible exchange (the Appendix shows the two arrangements). Within each combination of MC gender and register the audio is identical across the two visual conditions, meaning the visual factor changes nothing but the composition of the frame, and within each register the congresswoman and the congressman speak identical words. The two scripts take the same position on the same topic—the state of women's healthcare—and are matched in greeting, structure, and length, but they differ in emotional intensity and in how they close: the measured script ends on a solutions-oriented commitment, whereas the outraged script—which incorporates verbatim the emotionally charged on-screen text superimposed on the stimuli of that pilot study—ends in defiance (see Section 6.2 of the Appendix for every condition's full script).

Including rolling transcripts with the segments is deliberate: short news videos are primarily seen on platforms where they are often watched without sound \citep{kalogeropoulos2018newsvideo}. The condition-consistent stills respondents saw while completing the outcome measures carried the transcript. Participants were told at the outset that they would watch a short news clip created for this study, and a closing debrief disclosed that the MC was fictional and the voices were synthesized.

Respondents (N = 3,601 analyzed, 50\% women) completed the roughly five-minute survey on July 27, 2026, receiving \$1.50 for their participation (see the Appendix for sample demographics). Because the register manipulation is carried largely by the voice, a pre-treatment audio check preceded random assignment, screening out the fewer than one percent of entrants who could not hear a test clip, and a timer prevented advancing before the segment had played through. The four congresswoman conditions were each allocated 15 percent of the sample and the four congressman conditions 10 percent, an unequal allocation that concentrates power on the within-congresswoman contrasts central to our first and third hypotheses while preserving the full crossing. The registered stopping rule targeted 3,600 analyzed responses, a threshold fixed in advance by a pre-registered simulation, and collection closed one response past it, at 3,601. At that size the design afforded 98 percent power to detect a split-screen penalty of the size estimated in our first experiment; power for the registered interactions is detailed in the pre-analysis plan. Of 3,845 randomized completed responses, the primary analyzed sample retains the 3,601 (93.7 percent) passing a pre-registered instructed-response attention check. All confirmatory estimates are additionally reported with no attention exclusion (N = 3,845) and retaining only first-attempt audio passers (N = 3,594) in the Appendix, and no conclusion reported below depends on the choice among the three regimes.

After the video, participants completed the same outcome batteries as before—the policy-capability matrix from which we construct benchmarked perceived health expertise, and the three-item external efficacy battery ($\alpha = 0.74$). Every item referred to the MC gender-neutrally, and a condition-consistent still of the assigned segment, green box included, was re-displayed throughout the outcome block, exactly as in our first experiment. Two additions instrument the first causal link of each account developed in our theoretical expectations: for contested advocacy, a four-item battery recording perceived contestation; for agentic backlash, batteries recording perceived warmth (two items, $\alpha = 0.83$) and perceived norm violation (three items, $\alpha = 0.75$).

Following our pre-analysis plan, we report the two manipulation checks first; because the condition-consistent still remained visible while respondents answered, both were registered as checks that the manipulations were perceived, and neither gates any analysis. The outraged register raised perceived emotional intensity by nearly two points on the seven-point check ($\beta = 1.850$, $p < .001$), and respondents overwhelmingly identified their assigned format correctly: 96.5 percent of those assigned a solo condition reported the MC appeared alone, and 93.3 percent of those assigned a split-screen reported the MC appeared alongside another person.

\begin{figure}
     \caption{The congresswoman's split-screen penalty disappears when the broadcast appearance unfolds as dynamic video}
     \centering
     \includegraphics[width=\linewidth]{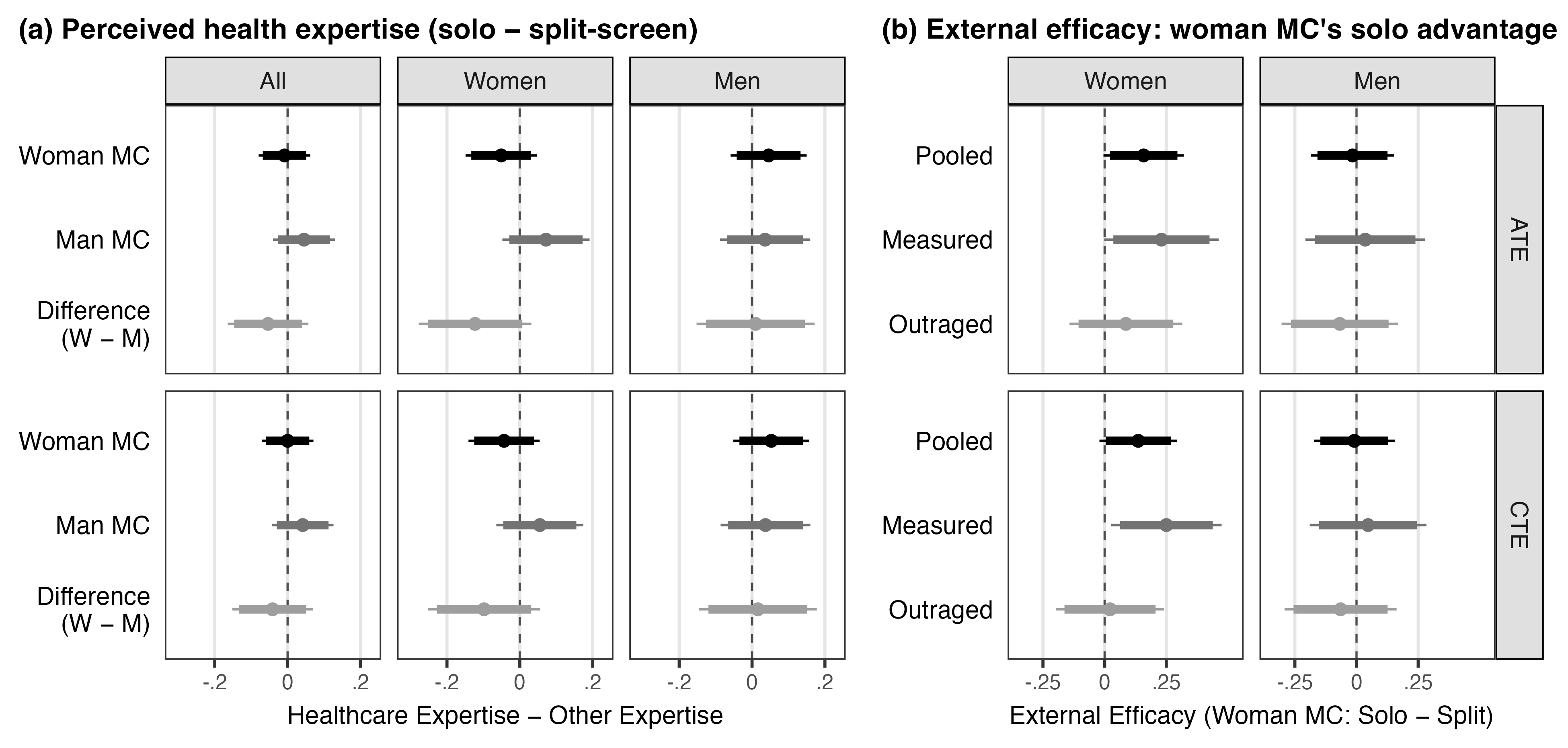}
     \floatfoot{\setstretch{1}\emph{Note}: Specification, controls, and display conventions follow Figure~\ref{fig:fig7}; both panels use Experiment 2's eight treatment cells, displayed as solo minus split-screen, so positive values indicate an advantage when appearing alone (in panel (a) this reverses the sign of the split-versus-solo coefficients reported in the text; panel (b) estimates are reported in the text in this orientation). Facet columns are respondent subgroups and contrasts are labeled on each panel's axis, with the woman MC's contrast in panel (a) and the pooled contrast in panel (b) set in black. Panel (a): N$_{all}=3601$, N$_{Women}=1786$, N$_{Men}=1765$ (3594, 1784, 1761 with covariates); tests are two-sided except the woman MC's contrast, registered as one-sided. Panel (b): the congresswoman's solo advantage in external efficacy among the women and men samples of panel (a), within each register and pooled over both; the pooled contrast among women respondents is the registered test of our second hypothesis (numbered H4 in our pre-analysis plan), and the register-specific contrasts are exploratory, within a family of roughly thirty subgroup comparisons (see the text); panel (b) tests are two-sided. Full results for both panels are in the Appendix.}
     \label{fig:fig11}
\end{figure}

With both manipulations perceived, the black lines in the first column of Figure \ref{fig:fig11}a report the dynamic analogue of our first hypothesis: the congresswoman's split-screen versus solo difference in perceived expertise, pooled over the two registers. We find no evidence of the penalty documented in the static designs. The difference is essentially zero ($\beta = 0.009$, 95\% CI $[-0.062, 0.079]$; registered one-sided $p_{\text{1s}} = .594$, two-sided $p = .813$), and the covariate-adjusted estimate is zero to three decimals ($\beta = -0.000$, 95\% CI $[-0.071, 0.071]$, $p = .997$). The null is also precisely estimated: the 90\% confidence interval, $[-0.051, 0.068]$, excludes not only a split-screen penalty of the size estimated in our first experiment ($\beta = 0.122$) but any penalty larger than half that size. The congressman's contrast is similarly null ($\beta = -0.045$, 95\% CI $[-0.130, 0.040]$, $p = .300$).\footnote{The rolling transcript these segments carry cannot itself explain the disappearance: in the static pilot study reported in the Appendix, the penalty persisted with substantive on-screen text present.}

The outraged register does not restore the penalty. Under the contested-advocacy account, our third hypothesis, the split-screen's cost should have concentrated where visual contest and outraged language coincide. Instead, the registered visual-by-register interaction for the congresswoman (oriented so that contested advocacy predicts a positive value) is wrong-signed and null ($\beta = -0.030$, 90\% CI $[-0.149, 0.089]$, $p_{\text{1s}} = .663$; two-sided $p = .675$), and is essentially unchanged with covariates ($\beta = -0.016$, $p_{\text{1s}} = .587$; two-sided $p = .825$). The outraged register cost the congresswoman no more in the split-screen ($\beta = -0.043$, 95\% CI $[-0.144, 0.057]$) than alone ($\beta = -0.074$, 95\% CI $[-0.174, 0.027]$).

The evidence for our fourth hypothesis is different. On perceived competence, the gender-by-register interaction is directionally consistent with agentic backlash but not statistically significant ($\beta = -0.067$, 95\% CI $[-0.178, 0.044]$, $p = .236$). On perceived warmth, the outcome on which the hypothesis was registered as confirmatory, the interaction is significant: identical outraged words cost the congresswoman more warmth than the congressman ($\beta = -0.268$, 95\% CI $[-0.448, -0.087]$, $p = .004$), and the estimate survives covariate adjustment ($\beta = -0.265$, 95\% CI $[-0.440, -0.090]$, $p = .003$). In the decomposition, the outraged register reduced the congresswoman's perceived warmth by $\beta = -0.415$ (95\% CI $[-0.522, -0.308]$, $p < .001$) and the congressman's by $\beta = -0.147$ (95\% CI $[-0.292, -0.003]$, $p = .046$), meaning the same words cost her roughly three times as much. Our first experiment isolated the double standard; this result is the double bind described in Section \ref{sec:review}.

The registered diagnostic for deciding between the two accounts is the congresswoman who expresses outrage while appearing alone: contested advocacy predicts no penalty in this condition, whereas agentic backlash predicts one. The effect of outrage in her solo condition is negative but not statistically significant ($\beta = -0.074$, 95\% CI $[-0.174, 0.027]$, $p = .150$; with covariates, $\beta = -0.068$, $p = .184$), and the registered equivalence test cannot settle the matter: the data rule out any benefit of outrage in this condition larger than our smallest effect size of interest, but they cannot rule out a penalty of that size.\footnote{The smallest effect size of interest, $d = 0.15$, was pre-registered and fixed while blind to all outcomes; full equivalence-test details for this and every registered bound are reported in the Appendix.}
The diagnostic is accordingly inconclusive, though the asymmetry (a benefit is ruled out, a penalty is not) leaves more room for backlash than for contested advocacy.

The perceptions we measured alongside these outcomes indicate why the comparison favors one account. Contested advocacy requires, as its first link, that the split-screen raise perceived contestation. It did not. If anything, respondents perceived slightly less contestation in the split-screen than in the solo shot ($\beta = -0.085$, $p = .033$), and the outraged register did nothing to change that.\footnote{The four-item contestation battery is the least reliable of our measures ($\alpha = 0.44$), and treatment effects on it should be read with corresponding caution.} The words were another matter. The outraged register increased perceived contestation by nearly half a scale point ($\beta = 0.434$, $p < .001$), for the congresswoman and the congressman alike. In short, viewers heard the contest in the language rather than seeing it in the frame.

Agentic backlash instead requires that outraged words depress perceived warmth. They did, as reported above, while the arrangement of the screen had no effect on warmth ($\beta = 0.012$, $p = .827$); outrage also raised perceived norm violation by a full scale point, again more for her than for him (both $p < .001$). Exploratory causal-mediation estimates, reported in the Appendix with the sensitivity analyses their assumptions require, point the same way: outrage's indirect effects on competence through warmth (ACME $= -0.038$) and norm violation (ACME $= -0.082$) are both negative ($p < .001$), while the split-screen's indirect effect through contestation is positive (ACME $= 0.010$), the opposite of the sign contested advocacy requires. Full coefficients and confidence intervals for every estimate in this section are provided in the Appendix.

We turn finally to our second hypothesis. As shown by the black lines in the first column of Figure \ref{fig:fig11}b, which pool over the two registers, women respondents who saw the congresswoman appear alone on screen reported higher external efficacy than those who saw her share it, an estimate that for the first time falls just short of statistical significance ($\beta = 0.158$, 95\% CI $[-0.005, 0.321]$, $p = .057$). The estimate attenuates somewhat under covariate adjustment ($\beta = 0.136$, 95\% CI $[-0.021, 0.293]$, $p = .090$), with all three items leaning the same way (see the Appendix). The registered equivalence test is again inconclusive, this time from the opposite side: the data rule out all but a small negative effect, but they cannot confirm that the positive movement is as large as our smallest effect size of interest.\footnote{As with the previous bound, the smallest effect size of interest ($d = 0.15$ on the efficacy scale) was pre-registered and fixed while blind to all outcomes, with full details in the Appendix.} 

The registered estimate pools over the two registers; panel (b) of Figure \ref{fig:fig11} separates the congresswoman's solo advantage by register and respondent gender, a moderation our pre-analysis plan designated as exploratory. Among women respondents, the solo advantage is carried almost entirely by the measured register ($\beta = 0.230$, 95\% CI $[-0.002, 0.463]$, $p = .052$; with the registered covariates, which include pre-treatment internal efficacy, $\beta = 0.250$, 95\% CI $[0.026, 0.474]$, $p = .028$). The corresponding advantage under the outraged register is small and far from significance ($\beta = 0.086$, 95\% CI $[-0.142, 0.315]$, $p = .460$). Among men respondents, every estimate is flat: the pooled solo advantage our plan treats as a descriptive companion to the registered test is essentially zero ($\beta = -0.016$, 95\% CI $[-0.186, 0.153]$, $p = .849$), neither register-specific advantage is distinguishable from zero, and no contrast comes closer to significance than $p = .143$.

Three findings from this experiment stand out. First, the dynamic split-screen carried no detectable competence cost for the congresswoman; the estimate is precise enough to exclude even half the static penalty, and the disappearance is most complete among the women respondents who drove that penalty, whose evaluations are essentially flat across her four conditions (see the Appendix). Second, the cost came from the words rather than the frame. Identical outraged language reduced the congresswoman's warmth roughly three times as much as the congressman's, the double bind our fourth hypothesis anticipated, while the contested-advocacy account failed at every measured step. Third, the movement in women viewers' external efficacy appears where a symbolic account would predict it, among women watching the congresswoman appear alone, speaking in the measured register whose closing lines describe a government that can solve problems. That pattern is exploratory. Of the roughly thirty covariate-adjusted subgroup contrasts of this kind, the measured-register estimate is the only one to cross the conventional threshold, and its unadjusted counterpart does not. Even so, it is the first movement this outcome has shown in any of our designs. All three conclusions are stable across the three registered exclusion regimes; full regime-by-regime tables and per-cell means and standard deviations for every condition and outcome are in the Appendix.

Set against Figure \ref{fig:fig5}, these experiments indicate when the disproportionate framing documented there should matter and when it should not. When an appearance is watched in full, it does a congresswoman little harm, because viewers judge the performance rather than the frame. When it is encountered as a static slice, the frame is nearly all the information available, and gendered expectations fill in the contest it implies, at the congresswoman's expense. The disproportionate assignment of congresswomen to split-screen shots is therefore best understood not as an invariant penalty but as an ongoing exposure: it feeds the circulating, stripped-down forms of the broadcast where the static penalty operates, while denying congresswomen the kind of appearance our exploratory evidence ties to women viewers' sense that government can answer to people like them.

\section{Discussion and Conclusion}

The contribution of this study is to specify when the visual framing of congresswomen's television appearances matters, for what, and for whom. Within the 695,464 minute-long segments of CNN and Fox News coverage that our women-related search terms retrieved, congresswomen appeared in 51.6 percent of split-screen frames but 32.6 percent of solo frames, which leaves them overrepresented, relative to their male colleagues, in the format that implies contest. Even in coverage nominally about women, a congresswoman who appears in a split-screen most often does so beside a congressman. What this skew costs depends on the form in which an appearance is encountered. When the appearance was a static frame, the split-screen imposed a competence penalty that emerged in every static design we fielded and that identically framed congressmen did not detectably share, consistent with the registered directional expectation that the difference is gendered. When the appearance unfolded as an exchange, the same composition had no measurable effect on the congresswoman's perceived competence, but the words carried a cost of their own: the outraged register reduced her warmth far more than the congressman's, however the screen was arranged. The two costs are analytically distinct. The split-screen penalty is a double standard that depends on the form of the encounter; the warmth penalty is a double bind that follows the speaker.

Read together, the experiments supply the scope conditions the visual-framing literature has largely lacked. Competence judgments depend on whatever evidence an encounter supplies. When viewers can watch a performance, the frame matters little; when all they have is a frozen frame, stereotype-based inference fills the gap. In the terms of our three-stage framework, this locates the harm at the interpretation stage rather than in exposure or framing itself. Two cautions bound this claim. Strictly speaking, our data show that the static penalty does not carry to dynamic anchor interviews; the medium and the split-screen partner changed together across our experiments, so the informational account is our interpretation of the difference. Our stimuli, moreover, were constructed rather than captured (fictional MCs and, in the dynamic experiment, synthesized voices, all disclosed to participants), which gave us full experimental control at some cost in realism, not least because real MCs come with partisanship, incumbency, and records attached, inviting motivated evaluations our deliberately nonpartisan stimuli do not engage.

Even so, the account organizes findings that otherwise sit in tension. The debate studies that anchor this literature found split-screen effects in fully dynamic footage \citep{cho2009split,stewart2017visual}; there, however, the adversary sharing the frame supplies exactly the contest the format implies, so watching confirms what the frame suggests, whereas our anchor interview corrects it. The same logic yields a falsifiable prediction: a dynamic split-screen pairing a congresswoman with a hostile adversary should revive the penalty. The account should also generalize. Compositional cues should matter most wherever performance is least observable, for other stereotype-marked politicians, other cues, and other information-poor encounters. This has a methodological implication as well. Because the penalty appears when the stimulus is a static frame but not when it is a video, the format of the stimulus is part of the treatment itself, not an incidental design choice.

That boundary condition would matter little if televised politics were always watched from start to finish. Increasingly, it is not. The penalty may be rare during the broadcast itself, but the conditions for it are re-created each time the appearance is clipped, muted, and screenshotted. Television remains among the most-cited individual sources of political news, but its audience skews older, and more Americans now encounter news through social and video platforms than through television itself; the majority of online news video is watched inside third-party platforms rather than on news organizations' own sites \citep{newman2025dnr}, and the news video that succeeds there is engineered to be watched without sound \citep{kalogeropoulos2018newsvideo}. Much of what circulates is not video at all: screen captures of television figure prominently among the most widely shared political images on social platforms, and authentic visuals traveling detached from their original context are the modal form of visual political content flagged by fact-checkers \citep{matatov2022stop, brennen2021beyond}. These are the information-poor encounters our literature review describes, and they are the conditions our static experiments simulate. The objection that a static frame is an unrealistic stimulus therefore gets the media environment backwards; for a growing share of the audience the still is the rule, and the attentive viewing that dissolved the penalty in our second experiment is the exception. We offer the implication as interpretation, since our data do not follow appearances onto platforms. But the roughly 27,000 split-screen frames containing a congresswoman in our corpus are candidate thumbnails, each carrying the frame's implication of contest into settings where nothing corrects it. The disproportionate split-screen assignment we document is thus compounded once these appearances leave the air.

The dynamic medium that protects a congresswoman's competence does not protect her warmth. Assigning the outraged register at random depressed perceived warmth for both speakers, but far more steeply for her: the gender-by-register interaction is the registered, confirmatory quantity ($\beta = -0.268$), and the decomposition it summarizes ($-0.415$ for the congresswoman against $-0.147$ for the congressman, both significant) is what ``roughly three times'' describes. The arrangement of the screen neither produced nor detectably altered this cost (full models in the Appendix). One caution attaches to this result. Because the statement concerned women, the design bundles the gender of the speaker with in-group advocacy (the congresswoman speaks about her own group where the congressman does not), so the gendered warmth cost may partly reflect that combination, a confound future designs could break by varying the group the statement concerns. Here the corpus and the experiment inform each other. Theory predicts that outrage is the register congresswomen bring to women's issues, yet our lexicon analysis found that register vanishingly rare in the coverage itself. The warmth result points to an answer. The language that would mark a congresswoman as a forceful advocate is also the language that costs her the most with viewers.

The second dimension distinguished at the outset is whether seeing a congresswoman in a position of authority leads the women watching to feel more capable themselves. On this dimension the confirmatory result is the null. In our first experiment, a momentary encounter did not move women viewers' external efficacy, even when the congresswoman commanded the screen alone, and the pre-registered equivalence test excludes all but small effects. Single static impressions, then, do not appreciably build efficacy, which is consistent with our account that orientations toward the political system rest on recognition; a still supplies a figure arranged within a shot rather than a person speaking to the viewer. The estimate approached conventional significance only in the dynamic experiment, where respondents watched and heard the congresswoman for forty seconds ($p = .057$). Even there, the exploratory breakdown concentrates the movement in a single configuration (women viewers watching the congresswoman alone in the measured register), and because that estimate is one of roughly thirty exploratory contrasts, we treat it as hypothesis-generating. Still, the pattern across the three outcomes is consistent: competence tracked the performance, warmth tracked the words, and efficacy moved, if it moved at all, with what the appearance signaled about the system's responsiveness, and only among the women watching.

If that exploratory pattern survives the test it still requires, its implication for the observational pattern in Figure \ref{fig:fig5} is direct. The configuration that moves women viewers pairs the register this coverage mostly supplies with the format in which congresswomen trail their male colleagues most. Measured speech is the norm in these broadcasts, where anger- and fear-laden vocabulary averages only about 29 words per 1,000. The solo shot is congresswomen's most common format in absolute terms, but it is also where they remain furthest from parity, with a men-to-women ratio of 2.1 in solo frames against 1.7 in split-screens. The same logic may help explain why findings on symbolic effects have been so persistently mixed. If the symbolic yield of visibility depends on its form and register rather than its volume alone, then studies measuring visibility as volume would find effects only where volume happened to arrive in the right form. We offer this as a testable hypothesis, and as a candidate explanation for the inconsistency with which our literature review began.

Future research should investigate the editorial and production logics that assign split-screen formats disproportionately to congresswomen relative to their male colleagues. Determining whether the prevalence of split-screen shots stems from conflict-driven news values, gendered booking practices, or broader visual conventions in political journalism matters because each origin points to a distinct institutional remedy. The circulation account opens a parallel agenda downstream of the broadcast, where the questions are which appearances get clipped and in what form they recirculate. On the reception side, because our stimuli carried on-screen text in every condition (from the headline text of the static frames to the rolling transcript of the dynamic segments), our design cannot identify what the text itself contributes; an experiment crossing viewer-selected captioning with the format of the shot would supply that answer. The measurement tools introduced here (pairwise facial recognition that identifies all 953 members of six Congresses from minimal training data per member, and zero-shot classification of shot composition) make this program tractable at the scale of the coverage itself.

Mediated visibility, \citet{brighenti2007} reminds us, is inherently ambivalent, capable of empowering and constraining at the same time. For congresswomen, we find, that ambivalence has a structure: their appearances are ample but disproportionately framed for contest; the frame's cost is paid where the appearance is encountered rather than watched; and forceful advocacy costs them warmth wherever it is heard. Little of this is within congresswomen's control, and none of it is theirs to fix. Advising women to temper their register would simply restate the double bind; the remedy lies instead with the editorial decisions that assign the split-screen and the platform defaults that decide how much of a performance survives. In a media environment where a broadcast minute outlives its broadcast, the double-edged sword of mediated visibility cuts less during the appearance itself than in what circulates afterward---the thumbnails, screenshots, and muted clips in which the frame is all that remains of the performance.

\begin{singlespace}
\bibliography{sample,press_coverage,references,congress_tv}
\end{singlespace}

\clearpage


\section*{Supplementary material}

# Online Appendix

## Double-Edged Sword of Mediated Visibility: How Visual Framing Undermines Congresswomen's Perceived Competence

# 1 The Siamese Neural Network

A Siamese Neural Network (SNN) is a type of neural network architecture that aims to determine similarity or dissimilarity between pairs of inputs. Instead of classifying a single input, an SNN considers pairs of data as inputs and predicts whether they are related in some meaningful way.

A SNN consists of two identical subnetworks that share architecture, parameters, and weights (the "Siamese" twin-like structure that gives the approach its name). Each subnetwork transforms one member of the input pair into an embedding vector capturing its essential characteristics, and the two embeddings are compared with a distance function. The model is trained with a loss function that pulls similar pairs together and pushes dissimilar pairs apart. In our paper, we used Euclidean distance and a contrastive loss.

Figure 1: Architecture of the Siamese Neural Network, with a worked example of its output

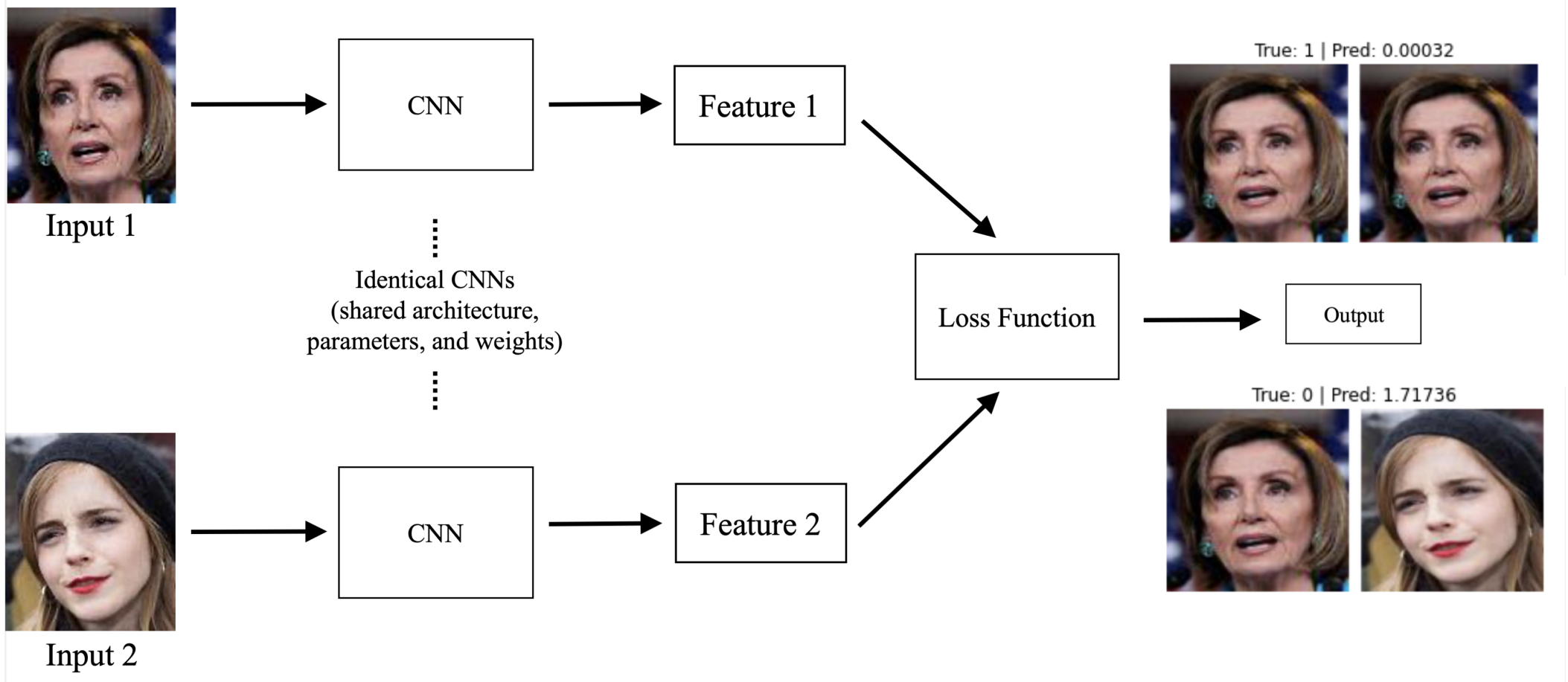


*Note*: Flowchart of the SNN used in this study, with example output at the far right. The number above each example pair is the predicted contrastive loss, with higher values implying less similarity; true pairs are labeled "True: 1," false pairs "True: 0." In the example, an image of Nancy Pelosi compared with herself yields a loss near zero (0.00032), while the same image compared with one of the actress Emma Watson yields a high value (1.71736), correctly separating two faces that share head position and facial structure.

Figure 1 shows the architecture of the network we trained, together with a worked example of its output. We trained the model for 500 epochs on our university's high-performance computing cluster, using one GPU node. Training took approximately eight hours.

Both the training and validation losses fell rapidly during the early epochs, with the training loss reaching approximately 0.02. The validation loss fluctuated between roughly 0.05 and 0.07 across the remaining epochs. This instability reflects the sensitivity of contrastive loss to pair sampling and margin saturation rather than conventional overfitting, since most negative pairs eventually satisfy the margin constraint and stop contributing to the loss.

## 2 The MC-versus-Celebrity Calculation

We obtained facial data of Members of Congress from Congress.gov, collecting official biographical images of individuals who served in the 112th through 117th Congresses. This resulted in a total of 953 legislators. We also used 1,100 celebrity images from a repository compiled by Mehta (2021). Because using all 2,053 images was infeasible given the enormous number of pairwise comparisons required, we applied the pre-processing steps below to construct the image pairs.

To make a feasible number of pairs while ensuring the model learns effectively, we subset the data by considering the facial orientation. We used the Google Cloud Vision API to identify pan, tilt, and roll information on top of each face's facial bounding boxes. Together, these three rotations (around the y-, x-, and z-axes, respectively) characterize the head orientation of each face.

$$\text{Distance} = \sqrt{(\text{pan}_{\text{legis}} - \text{pan}_{\text{celeb}})^2 + (\text{tilt}_{\text{legis}} - \text{tilt}_{\text{celeb}})^2 + (\text{roll}_{\text{legis}} - \text{roll}_{\text{celeb}})^2} \quad (1)$$

To control for facial pose differences and reduce noise in training, we prioritized pairing faces with similar orientation. Using the pan, tilt, and roll values, we calculated the three-dimensional Euclidean distance (see Equation 1) between each legislator and celebrity pair. A smaller distance indicates greater similarity in facial orientation, which helps the model focus on learning identity-relevant features rather than being confounded by pose variations.

For each legislator, we identified the 20 celebrities with the smallest distance values. As a concrete example, the actress Kate Bosworth is among the 20 celebrities whose facial orientation is most similar to that of Nancy Pelosi's official photo, with a facial-orientation distance of 2.1438. For Pelosi, the closest celebrity has a distance of 1.8271, while the farthest among the top 20 has 4.4203. These minimum and maximum distance values vary across different members of Congress, indicating individual-level variation in facial orientation similarity scores.

To determine whether a face extracted from our image data was a member of Congress (MC), we did the following:

1. Identify 20 celebrity images whose gender and facial orientation best matched the MC. The gender was perfectly matched. Facial orientation was matched using a weighted random sample in which celebrity faces whose facial distance (see Equation 1) to the MC was smallest were given a higher probability of being selected.
2. For each face extracted from a frame from our dataset, we calculated the facial similarity between that face and the MC's face from their Congress.gov photo. This was done using the SNN described in Section 1 and yielded a single facial similarity score ($SNN_{MC}$).
3. We then compared the same extracted face to the 20 celebrity images described in Step 1 using the SNN described in Section 1, yielding a vector of facial similarity scores ($SNN_{Celebrity1}, SNN_{Celebrity2}, \ldots, SNN_{Celebrity20}$). Using these 20 facial similarity scores, we created a 95% confidence interval using the mean and $\pm 1.96$ standard deviations.
4. If $SNN_{MC}$ was higher than the upper bounds of the 95% confidence interval described

in Step 3, then we said the face extracted from the frame in our dataset was the MC's face.

# 3 Validation of the Facial Recognition Pipeline

## 3.1 Emma Watson Calculation

To validate our facial recognition approach, we replicated the Nancy Pelosi–Emma Watson calculation presented in Figure 1. This included the following steps:

1. Using the SNN described in Section 1, we compared an MC's Congress.gov photo to itself, yielding a facial similarity score.
2. We then compared the MC's Congress.gov photo to 20 randomly selected celebrities who share the same gender and facial orientation, yielding a vector of facial similarity scores.
3. Steps 1 and 2 were repeated for each MC and two means were calculated. The first mean was for the results of Step 1. The second mean was for the results of Step 2.

The model prediction when the MC's face was compared to itself did not significantly differ from zero ($\bar{x} = 0.0003162278$), whereas the same cannot be said of the comparisons with the randomly selected celebrities ($\bar{x} = 2.12483$), suggesting the SNN reasonably differentiates an MC's face from the faces of celebrities who share the same gender and facial orientation.

## 3.2 Transcript-Based Calculation

We employed a text-based measure to establish convergent validity for the facial recognition approach we used in this study. First, we used a collection of more than 120,000 LexisNexis transcripts (2011–2021) from CNN and Fox News alongside one-minute closed-caption blocks from the Internet Archive to determine when MCs likely appeared on screen. Mentions of MCs were matched to the congress-legislators GitHub repository (unitedstates project, 13 ) and cross-checked against on-air introduction text to ensure we captured only currently serving members, excluding political candidates or those previously serving in other occupations or positions.

Closed-caption sections were truncated to 256 tokens and transcripts split into rolling windows of similar size. Both were encoded with a sentence-transformer model (all-MiniLM-L6-v2), and we retained the caption window with the highest cosine similarity when it exceeded a 0.85 threshold, determined through manual verification on a random sample of 100.

We then identified which MC was talking in each paired segment from the nearest preceding speaker identifier in the transcript (e.g., "PELOSI:"), and assumed that MC appeared in the paired frame whenever the frame contained a human face according to the Google Vision API.

Convergent validity involves showing that two instruments intended to measure the same construct yield similar results (Chin and Yao, 2024). This text-based measure reproduces the study's substantive results. When appearances are identified from transcripts rather than faces, congresswomen still appear on CNN and Fox News more often than their representation on Capitol Hill would predict, and still appear disproportionately in split-screen shots. This convergence indicates that the patterns in the main text are not artifacts of the

facial recognition pipeline. It does not, however, make the two approaches interchangeable, because a transcript cannot show whether a legislator was on screen or how the shot was composed.

# 4 Corpus Retrieval and Shot Classification

## 4.1 Per-Term Search Frequencies

As described in the Data and Methods section of the main text, the corpus was constructed by applying the sixteen woman-related terms of the Pearson and Dancey (2011) dictionary, adopted without modification, to CNN and Fox News broadcasts in the TV News Archive from May 13, 2011, to March 16, 2021. The dictionary's terms pick out coverage where women are the topic rather than naming particular persons, which is why "congresswoman" and "congresswomen" are absent while "servicewoman" and "servicewomen" appear. We adopted the published list wholesale so that no term reflects our own judgment. Table 1 reports the number of image-text segments each term returned on each network, with a segment counting toward a term when that exact form appears at least once in its paired minute of closed-caption text (possessive and plural variants counted separately, as the dictionary lists them). Because several terms often co-occur within a minute, the per-term counts sum to more than the 344,863 CNN and 294,192 Fox News segments containing at least one term. The remaining segments were returned by the Archive's search although no dictionary term appears in their own paired minute of captions.

Table 1: Number of image-text segments returned by each term in the Pearson and Dancey (2011) dictionary, May 13, 2011–March 16, 2021

| | CNN | Fox News | | CNN | Fox News |
|---|---|---|---|---|---|
| "woman" | 119,834 | 104,350 | "female" | 51,218 | 34,694 |
| "women" | 134,904 | 119,265 | "female's" | 40 | 30 |
| "woman's" | 4,967 | 5,002 | "females" | 1,276 | 1,291 |
| "women's" | 13,305 | 11,940 | "females'" | 2 | 2 |
| "girl" | 43,395 | 35,595 | "servicewoman" | 7 | 3 |
| "girl's" | 2,094 | 2,125 | "servicewoman's" | 0 | 0 |
| "girls" | 26,316 | 20,942 | "servicewomen" | 19 | 11 |
| "girls'" | 724 | 467 | "servicewomen's" | 1 | 1 |
| Segments matching any term | 344,863 | 294,192 | All segments | 374,643 | 320,821 |

*Note:* Cells count segments containing the term at least once.

## 4.2 Shot Classification with CLIP

To determine whether frames containing two legislators were presented as split-screen shots, we used CLIP, a vision–language model that maps images and natural-language prompts into a shared embedding space and is widely applied to "zero-shot" image classification (Radford et al., 2021). Rather than training a custom classifier, which would have required extensive manual coding of training examples, we constructed a small set of textual descriptions of prototypical shots (e.g., "a TV split-screen of two people" versus "a single camera shot of one scene") and computed the similarity between each frame and each prompt, converting the similarities into the probability that the frame is a split-screen shot. A frame was classified as

a split-screen when this probability exceeded a decision threshold of 0.70, which was specified before classification began. Applied to the frames in which the facial-recognition pipeline of Sections 1–3 identified exactly two legislators, this procedure classified 27,658 CNN frames and 25,519 Fox News frames as split-screen shots. These are the split-screen counts used throughout the main text.

# 5 Experiment 1: The Fully Crossed Static Design

## 5.1 Samples of the Two 2026 Experiments

Table 2 reports the demographic composition of the two Prolific samples used in Experiments 1 and 2 of the main text, alongside the four pilot samples reported in Section 8.

Table 2: Demographic breakdown of the six survey samples: the two 2026 experiments and the four pilot samples

| | *2026 Experiments* | | *2024 Pilot Waves* | | *Dec. 2024 Pilot + Repl.* | |
|---|---|---|---|---|---|---|
| | Exp. 1 | Exp. 2 | Qualtrics | Students | Prolific | MTurk |
| Dates | 7/23/26 | 7/27/26 | Feb. 2024 | Apr. 2024 | 12/10/24 | 5/15/25 |
| N | 2,619 | 3,601 | 925 | 195 | 500 | 130 |
| Democrat (%) | 44 | 45 | 40 | 46 | 38 | 51 |
| College Degree (%) | 56 | 56 | 49 | 10 | 65 | 99 |
| White (%) | 69 | 70 | 61 | 61 | 66 | 91 |
| Women (%) | 49 | 50 | 54 | 55 | 52 | 26 |
| Age (mean) | 42.0 | 41.3 | — | — | — | — |

*Note*: Experiment Ns are analyzed samples under each experiment's pre-registered exclusion regime (Experiment 1: 2,619 of 2,733 completes; Experiment 2: 3,601 of 3,845 eligible completes); the four pilot columns reproduce Table 1 of the pre-revision manuscript, which did not report sample ages. The pilot studies are reported in Section 8.

## 5.2 Stimuli and Vignette

Every respondent read the following vignette, presented with the assigned frame (the frame appeared above the text, and the continue button appeared after ten seconds):

> "Cable news has recently been covering proposals in Congress to support frontline healthcare providers. Recognizing the vital contributions of healthcare providers — especially women — the proposals focus on the distinct challenges they face in areas such as research, patient care, and work-life balance. Supporters say the goal is not just to level the playing field for women, but to give healthcare providers the tools and opportunities they need to drive innovation and improve the quality and effectiveness of care. Above is an image from one of these recent broadcasts, where [the members of Congress shown were / the member of Congress shown was] invited to discuss the proposals. Please review it carefully — the button to continue appears after 10 seconds."

The bracketed phrase was piped to match the assigned condition. Figure 2 reports the complete set of frames as fielded.

Figure 2: The six treatment frames of Experiment 1 as fielded, with an example evaluation-stage frame in the final panel

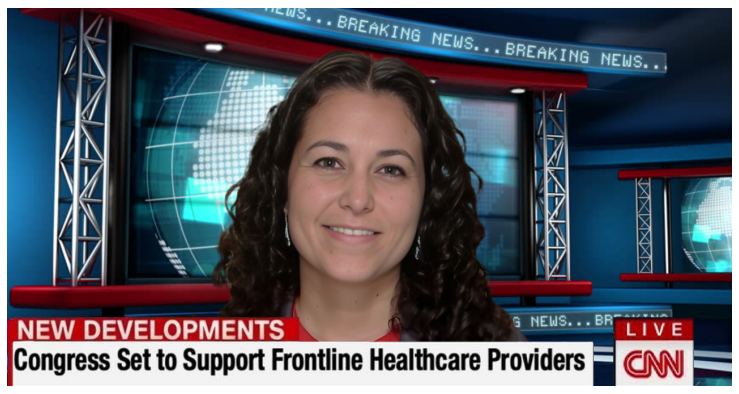


Woman MC, solo

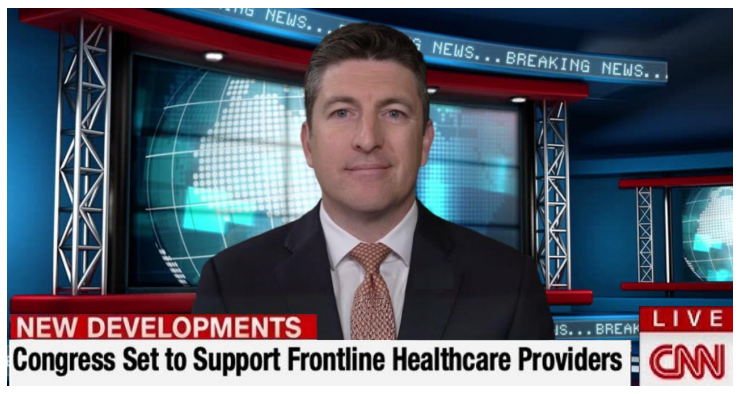


Man MC, solo

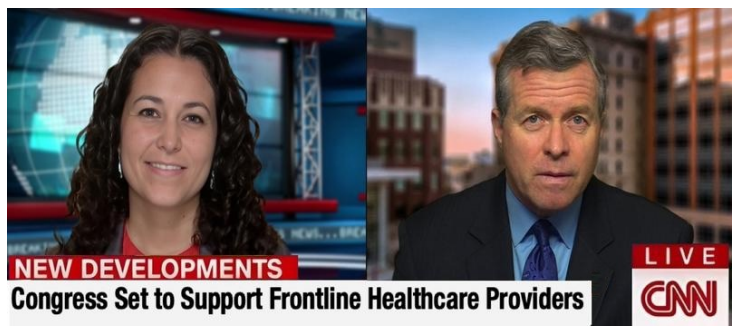


Woman MC, split-screen, man colleague

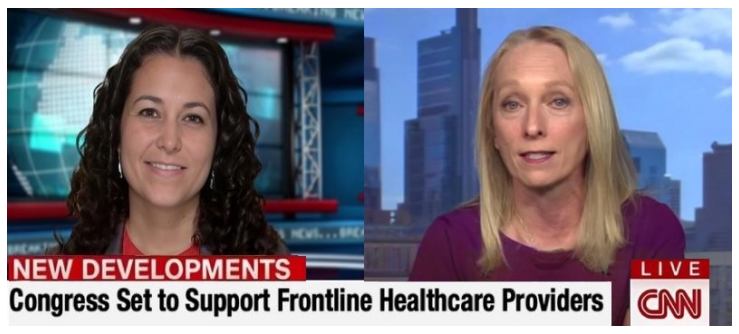


Woman MC, split-screen, woman colleague

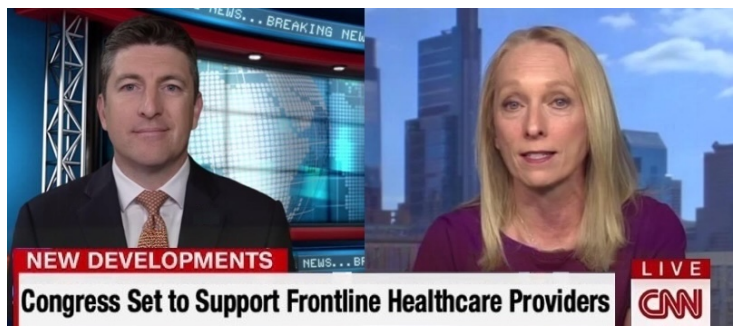


Man MC, split-screen, woman colleague

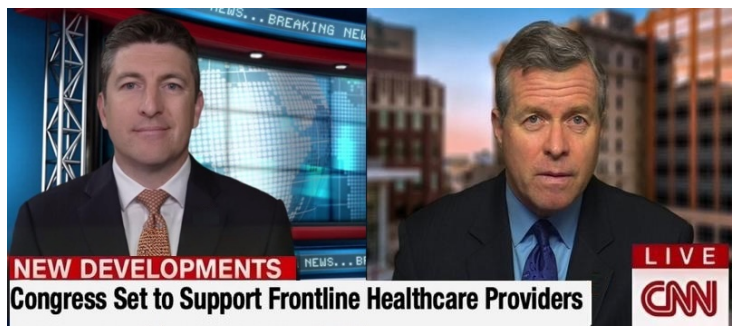


Man MC, split-screen, man colleague

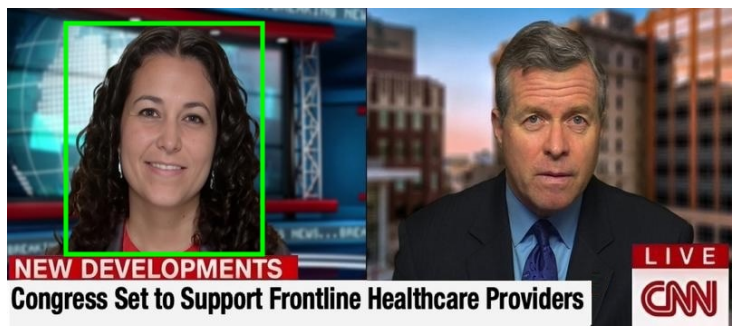


Evaluation stage (example)

*Note*: The six treatment frames as presented alongside the vignette text, with the evaluation-stage example in the final panel. Only the gender of the MC and the format of the shot vary across conditions; the two solo conditions were each allocated 25 percent of the sample and the four split-screen conditions 12.5 percent apiece.

## 5.3 The Complete Set of Treatment Frames

The six conditions were built from the same four individuals—a woman MC, a man MC, a woman colleague, and a man colleague—with identical graphics and accompanying text in every condition, and the chyron always read "Congress Set to Support Frontline Healthcare Providers." Figure 2 presents the six frames respondents saw alongside the vignette text. The final panel of Figure 2 shows an example of the corresponding evaluation stage, in which the assigned frame was re-displayed with a green box identifying the MC being evaluated and remained visible while respondents completed the outcome batteries. The other five evaluation-stage frames apply the same green box to their treatment frames. This is the same device used to mark the MC in Experiment 2.

## 5.4 Outcome Measures as Fielded

**Political capability.** "Please tell us how well do you think the member of Congress in green would be able to address the issues listed below": *Healthcare policies*, *National Security policies*, *Education policies*, *Economic policies*, each rated from "Poorly" to "Very well" on a five-point scale. As in the main text, the healthcare response was benchmarked against the mean evaluation of the three other policy areas.

**External efficacy.** "Please indicate the extent to which you agree or disagree with the following statements": *I feel most public officials care what people like me think*; *I feel people like me have a say in what the government does*; *I feel no matter whom I vote for, it won't make a difference to what happens in the country* (reverse-coded), each on a seven-point scale from "Strongly disagree" to "Strongly agree," coded −3 to 3 and averaged. Experiment 2 used the same two batteries and added the contestation, warmth, and norm-violation batteries described in the main text.

## 5.5 Full Results

Tables 3 and 4 report every estimand of Experiment 1's registered analysis under the primary exclusion regime, for both specifications and all three respondent groups. Estimands are named as in the study's analysis archive. Table 5 reports the registered quantities under all three registered attention regimes, and Table 6 the registered item-level external-efficacy estimates for both experiments. We conducted the equivalence tests following the pre-analysis plan. For the efficacy hypothesis among women respondents, we can rule out effects outside $|d| = 0.153$ at 90 percent confidence ($\beta = 0.031$, 90% CI $[-0.141, 0.203]$; index $\alpha = 0.74$). No equivalence threshold was registered for the man MC's penalty, so we report only its confidence interval; the largest $|d|$ consistent with that 90% CI is 0.121. Table 7 reports per-cell means and standard deviations for the full analyzed sample under every registered exclusion regime.

Table 3: Experiment 1, full coefficient estimates, primary registered exclusion regime: perceived health expertise (benchmarked)

| | ATE | | | | CTE | | | |
|---|---|---|---|---|---|---|---|---|
| Subgroup | $n$ | $\beta$ | 95% CI | $p$ | $n$ | $\beta$ | 95% CI | $p$ |
| *Woman MC: solo − split-screen (pooled over colleague gender)* (`female_penalty_pooled`) | | | | | | | | |
| All | 2,618 | 0.122 | [0.044, 0.199] | .002 | 2,601 | 0.130 | [0.053, 0.207] | .001 |
| Women | 1,295 | 0.140 | [0.037, 0.244] | .008 | 1,286 | 0.152 | [0.048, 0.257] | .004 |
| Men | 1,281 | 0.097 | [−0.023, 0.216] | .113 | 1,273 | 0.105 | [−0.012, 0.222] | .079 |
| *Man MC: solo − split-screen (pooled over colleague gender)* (`male_penalty_pooled`) | | | | | | | | |
| All | 2,618 | 0.026 | [−0.049, 0.101] | .501 | 2,601 | 0.025 | [−0.050, 0.101] | .509 |
| Women | 1,295 | 0.023 | [−0.086, 0.133] | .679 | 1,286 | 0.024 | [−0.086, 0.134] | .673 |
| Men | 1,281 | 0.026 | [−0.078, 0.130] | .624 | 1,273 | 0.031 | [−0.073, 0.135] | .557 |
| *Difference in penalties: woman MC − man MC (pooled)* (`interaction_pooled`) | | | | | | | | |
| All | 2,618 | 0.096 | [−0.012, 0.204] | .082 (.041) | 2,601 | 0.105 | [−0.003, 0.212] | .056 (.028) |
| Women | 1,295 | 0.117 | [−0.033, 0.268] | .127 | 1,286 | 0.129 | [−0.023, 0.280] | .096 |
| Men | 1,281 | 0.071 | [−0.088, 0.229] | .383 | 1,273 | 0.074 | [−0.083, 0.230] | .357 |
| *Woman MC: solo − split-screen beside a man colleague* (`female_penalty_mixed`) | | | | | | | | |
| All | 2,618 | 0.164 | [0.071, 0.258] | .001 | 2,601 | 0.181 | [0.090, 0.272] | $<$ .001 |
| Women | 1,295 | 0.165 | [0.041, 0.289] | .009 | 1,286 | 0.184 | [0.058, 0.309] | .004 |
| Men | 1,281 | 0.158 | [0.014, 0.302] | .031 | 1,273 | 0.173 | [0.037, 0.309] | .012 |
| *Difference in penalties, mixed-colleague contrast* (`interaction_mixed`) | | | | | | | | |
| All | 2,618 | 0.064 | [−0.067, 0.196] | .336 (.168) | 2,601 | 0.078 | [−0.051, 0.207] | .237 (.119) |
| Women | 1,295 | 0.007 | [−0.175, 0.190] | .936 | 1,286 | 0.017 | [−0.166, 0.201] | .854 |
| Men | 1,281 | 0.117 | [−0.076, 0.310] | .234 | 1,273 | 0.124 | [−0.063, 0.311] | .194 |
| *Woman MC: woman colleague − man colleague (registered exploratory)* (`partner_female_target`) | | | | | | | | |
| All | 2,618 | −0.085 | [−0.191, 0.021] | .116 | 2,601 | −0.101 | [−0.205, 0.003] | .057 |
| Women | 1,295 | −0.050 | [−0.193, 0.093] | .496 | 1,286 | −0.063 | [−0.206, 0.081] | .393 |
| Men | 1,281 | −0.123 | [−0.283, 0.037] | .131 | 1,273 | −0.137 | [−0.291, 0.017] | .081 |
| *Man MC: woman colleague − man colleague (registered exploratory)* (`partner_male_target`) | | | | | | | | |
| All | 2,618 | −0.148 | [−0.249, −0.047] | .004 | 2,601 | −0.155 | [−0.256, −0.053] | .003 |
| Women | 1,295 | −0.269 | [−0.413, −0.126] | $<$ .001 | 1,286 | −0.285 | [−0.430, −0.140] | $<$ .001 |
| Men | 1,281 | −0.030 | [−0.174, 0.114] | .682 | 1,273 | −0.036 | [−0.179, 0.107] | .625 |

Table 4: Experiment 1, full coefficient estimates, primary registered exclusion regime: external efficacy

| | ATE | | | | CTE | | | |
|---|---|---|---|---|---|---|---|---|
| Subgroup | $n$ | $\beta$ | 95% CI | $p$ | $n$ | $\beta$ | 95% CI | $p$ |
| *Woman MC: solo − split-screen (pooled over colleague gender)* (`female_penalty_pooled`) | | | | | | | | |
| All | 2,619 | 0.077 | [−0.073, 0.227] | .317 | 2,602 | 0.079 | [−0.067, 0.225] | .287 |
| Women | 1,295 | 0.031 | [−0.174, 0.236] | .768 | 1,286 | 0.033 | [−0.169, 0.234] | .751 |
| Men | 1,282 | 0.109 | [−0.117, 0.334] | .346 | 1,274 | 0.128 | [−0.091, 0.346] | .252 |
| *Man MC: solo − split-screen (pooled over colleague gender)* (`male_penalty_pooled`) | | | | | | | | |
| All | 2,619 | 0.035 | [−0.113, 0.182] | .643 | 2,602 | 0.057 | [−0.087, 0.201] | .438 |
| Women | 1,295 | 0.117 | [−0.088, 0.322] | .263 | 1,286 | 0.141 | [−0.057, 0.338] | .163 |
| Men | 1,282 | −0.055 | [−0.271, 0.161] | .619 | 1,274 | −0.029 | [−0.242, 0.185] | .792 |
| *Difference in penalties: woman MC − man MC (pooled)* (`interaction_pooled`) | | | | | | | | |
| All | 2,619 | 0.042 | [−0.169, 0.252] | .697 (.349) | 2,602 | 0.022 | [−0.183, 0.228] | .832 (.416) |
| Women | 1,295 | −0.086 | [−0.376, 0.204] | .561 | 1,286 | −0.108 | [−0.390, 0.174] | .453 |
| Men | 1,282 | 0.164 | [−0.149, 0.476] | .305 | 1,274 | 0.156 | [−0.149, 0.462] | .315 |
| *Woman MC: solo − split-screen beside a man colleague* (`female_penalty_mixed`) | | | | | | | | |
| All | 2,619 | 0.134 | [−0.051, 0.318] | .156 | 2,602 | 0.141 | [−0.039, 0.320] | .125 |
| Women | 1,295 | 0.073 | [−0.180, 0.326] | .572 | 1,286 | 0.060 | [−0.190, 0.311] | .636 |
| Men | 1,282 | 0.182 | [−0.093, 0.458] | .194 | 1,274 | 0.228 | [−0.037, 0.493] | .092 |
| *Difference in penalties, mixed-colleague contrast* (`interaction_mixed`) | | | | | | | | |
| All | 2,619 | 0.063 | [−0.197, 0.322] | .636 (.318) | 2,602 | 0.039 | [−0.213, 0.291] | .761 (.380) |
| Women | 1,295 | −0.077 | [−0.439, 0.285] | .677 | 1,286 | −0.136 | [−0.487, 0.214] | .446 |
| Men | 1,282 | 0.179 | [−0.200, 0.558] | .355 | 1,274 | 0.201 | [−0.168, 0.569] | .286 |
| *Woman MC: woman colleague − man colleague (registered exploratory)* (`partner_female_target`) | | | | | | | | |
| All | 2,619 | −0.114 | [−0.328, 0.100] | .296 | 2,602 | −0.123 | [−0.329, 0.084] | .245 |
| Women | 1,295 | −0.084 | [−0.378, 0.210] | .576 | 1,286 | −0.056 | [−0.345, 0.233] | .706 |
| Men | 1,282 | −0.148 | [−0.466, 0.170] | .362 | 1,274 | −0.201 | [−0.504, 0.103] | .196 |
| *Man MC: woman colleague − man colleague (registered exploratory)* (`partner_male_target`) | | | | | | | | |
| All | 2,619 | −0.072 | [−0.286, 0.141] | .507 | 2,602 | −0.089 | [−0.298, 0.121] | .406 |
| Women | 1,295 | −0.066 | [−0.354, 0.222] | .653 | 1,286 | −0.112 | [−0.390, 0.166] | .429 |
| Men | 1,282 | −0.117 | [−0.439, 0.204] | .475 | 1,274 | −0.112 | [−0.430, 0.206] | .489 |

Table 5: Experiment 1, registered quantities under all three registered attention regimes (ATE), with $n$'s as per-outcome complete cases

| Quantity | Regime | $n$ | $\beta$ | 95% CI | $p$ |
|---|---|---|---|---|---|
| Woman MC solo − split, competence (H1) | Primary (revised rule) | 2,618 | 0.122 | [0.044, 0.199] | .002 |
| | Strict rule | 1,960 | 0.138 | [0.052, 0.224] | .002 |
| | No exclusion | 2,730 | 0.132 | [0.056, 0.209] | .001 |
| Man MC solo − split, competence | Primary (revised rule) | 2,618 | 0.026 | [−0.049, 0.101] | .501 |
| | Strict rule | 1,960 | 0.039 | [−0.047, 0.124] | .377 |
| | No exclusion | 2,730 | 0.032 | [−0.041, 0.105] | .387 |
| Difference in penalties (registered one-sided) | Primary (revised rule) | 2,618 | 0.096 | [−0.012, 0.204] | .082 (.041) |
| | Strict rule | 1,960 | 0.099 | [−0.022, 0.220] | .108 (.054) |
| | No exclusion | 2,730 | 0.100 | [−0.005, 0.206] | .063 (.032) |
| Woman MC solo − split, efficacy, women (H2) | Primary (revised rule) | 1,295 | 0.031 | [−0.174, 0.236] | .768 |
| | Strict rule | 994 | 0.076 | [−0.160, 0.311] | .528 |
| | No exclusion | 1,341 | 0.025 | [−0.176, 0.227] | .804 |

# 6 Experiment 2: The Dynamic Design

## 6.1 Stimulus Frames

Figure 3 shows the two visual arrangements of Experiment 2's segments: the solo condition, in which the screen cuts to whoever is speaking, and the split-screen condition, in which the MC and the anchor share a two-shot throughout.

## 6.2 Full Scripts for the Segments

Every segment follows the same three-turn structure: the anchor's introduction, the MC's statement, and the anchor's close. The rolling transcript displayed these words phrase by

Table 6: Item-level external-efficacy estimates for the woman MC's solo-versus-split-screen contrast, women respondents, primary regimes

| | ATE | | | CTE | | |
|---|---|---|---|---|---|---|
| Outcome | $\beta$ | 95% CI | $p$ | $\beta$ | 95% CI | $p$ |
| *Experiment 1: solo versus split-screen, woman MC, women respondents* | | | | | | |
| External efficacy index | 0.031 | [−0.174, 0.236] | .768 | 0.033 | [−0.169, 0.234] | .751 |
| Officials care what people like me think | 0.045 | [−0.187, 0.277] | .704 | 0.088 | [−0.144, 0.320] | .457 |
| People like me have a say | 0.039 | [−0.216, 0.294] | .764 | 0.035 | [−0.217, 0.286] | .786 |
| My vote makes no difference (raw coding) | −0.009 | [−0.284, 0.266] | .950 | 0.025 | [−0.244, 0.294] | .854 |
| *Experiment 2: solo versus split-screen, woman MC, women respondents* | | | | | | |
| External efficacy index | 0.158 | [−0.005, 0.321] | .057 | 0.136 | [−0.021, 0.293] | .090 |
| Officials care what people like me think | 0.127 | [−0.055, 0.309] | .172 | 0.118 | [−0.063, 0.299] | .203 |
| People like me have a say | 0.139 | [−0.064, 0.341] | .179 | 0.130 | [−0.068, 0.327] | .198 |
| My vote makes no difference (reverse-coded) | 0.209 | [−0.004, 0.423] | .055 | 0.161 | [−0.043, 0.365] | .121 |

Table 7: Experiment 1, per-cell means and standard deviations, full analyzed sample, every registered exclusion regime, including the raw five-point healthcare item

| Regime | Condition | $n$ | Raw healthcare item | | Health expertise | | External efficacy | |
|---|---|---|---|---|---|---|---|---|
| | | | mean | SD | mean | SD | mean | SD |
| Primary (revised rule) | Woman MC, solo | 655 | 3.973 | 0.898 | 0.462 | 0.740 | −0.359 | 1.376 |
| | Woman MC, woman coll. | 325 | 3.828 | 0.896 | 0.383 | 0.700 | −0.378 | 1.390 |
| | Woman MC, man coll. | 333 | 3.784 | 0.985 | 0.297 | 0.690 | −0.492 | 1.410 |
| | Man MC, solo | 653 | 3.348 | 0.915 | 0.045 | 0.721 | −0.403 | 1.327 |
| | Man MC, man coll. | 328 | 3.378 | 0.934 | 0.094 | 0.636 | −0.401 | 1.393 |
| | Man MC, woman coll. | 325 | 3.240 | 0.955 | −0.054 | 0.681 | −0.474 | 1.394 |
| Strict rule | Woman MC, solo | 479 | 3.996 | 0.899 | 0.467 | 0.722 | −0.354 | 1.349 |
| | Woman MC, woman coll. | 251 | 3.805 | 0.911 | 0.359 | 0.663 | −0.385 | 1.379 |
| | Woman MC, man coll. | 247 | 3.822 | 0.920 | 0.300 | 0.621 | −0.501 | 1.395 |
| | Man MC, solo | 479 | 3.334 | 0.902 | 0.036 | 0.702 | −0.393 | 1.308 |
| | Man MC, man coll. | 252 | 3.349 | 0.930 | 0.062 | 0.632 | −0.456 | 1.380 |
| | Man MC, woman coll. | 253 | 3.241 | 0.960 | −0.067 | 0.695 | −0.414 | 1.399 |
| No exclusion | Woman MC, solo | 681 | 3.956 | 0.908 | 0.459 | 0.752 | −0.347 | 1.362 |
| | Woman MC, woman coll. | 341 | 3.809 | 0.886 | 0.364 | 0.696 | −0.374 | 1.387 |
| | Woman MC, man coll. | 346 | 3.769 | 0.989 | 0.289 | 0.696 | −0.446 | 1.414 |
| | Man MC, solo | 681 | 3.336 | 0.922 | 0.044 | 0.716 | −0.395 | 1.318 |
| | Man MC, man coll. | 345 | 3.362 | 0.924 | 0.081 | 0.627 | −0.393 | 1.379 |
| | Man MC, woman coll. | 337 | 3.234 | 0.949 | −0.057 | 0.673 | −0.449 | 1.395 |

phrase as the synthesized voices delivered them, so the scripts below are both the audio respondents heard and the text they saw. The anchor is the same man, with the same voice, in all eight conditions, and his introduction varies only in the fictional MC's first name. Within each register the congresswoman and the congressman speak word-for-word identical statements, and within each combination of MC gender and register the audio track is identical across the solo and split-screen conditions.

**Anchor introduction (all conditions; only the MC's first name varies).**

> "Turning now to healthcare. We're joined by Representative Diane [David] Hartwell of the House Energy and Commerce Committee. Representative, what's your read on where things stand for women right now?"

**MC statement, measured register (identical for the congresswoman and the congressman).**

> "Thanks for having me, Brian. There is a lot to cover. First, I am focused on the state of healthcare. Women often face challenges that are overlooked, unaccounted for, and ultimately unaddressed. And I'm committed to working on solutions."

Figure 3: Frames from the video segments used in Experiment 2, showing the woman MC in the solo condition (left) and sharing a two-shot with the anchor in the split-screen condition (right), with the rolling transcript visible in both

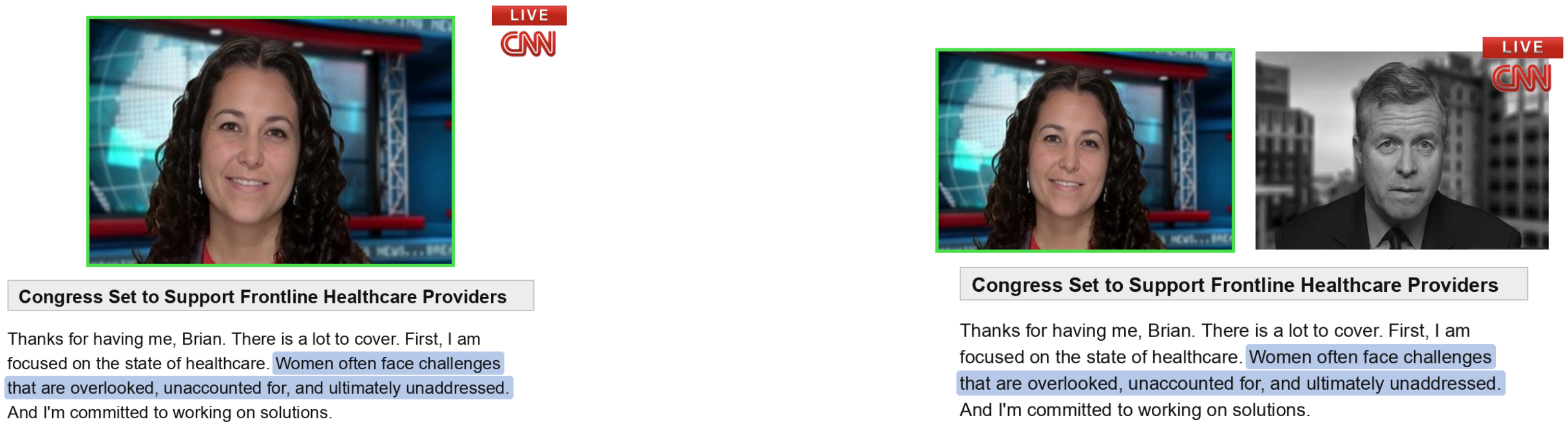


*Note*: Both frames are from the woman MC's measured-register segments at the same point in the statement; the green box marks the MC being evaluated, and the rolling transcript is visible at the bottom. In the split-screen conditions the active speaker appears in color and the listener in greyscale. These same stills were re-displayed while respondents completed the outcome batteries.

**MC statement, outraged register (identical for the congresswoman and the congressman).**

> "Thanks for having me, Brian. I don't know where to begin. First, I am anxious about the state of healthcare. Women are often left outraged, terrified, and forgotten. And I won't stay quiet while that happens."

**Anchor close (all conditions).**

> "We've got a lot more to get to here. Stay with me, Representative Hartwell while I take a quick break."

The two statements are matched in greeting, structure, and length (36 tokens for the outraged statement and 39 for the measured, under the tokenization of Section 7), take the same position on the same topic, and differ in emotional intensity and in closing stance. The measured script closes on a solutions-oriented commitment and the outraged script closes in defiance. The middle of the outraged statement reproduces verbatim the on-screen text superimposed on the stimuli of the December 2024 pilot (Section 8). Section 7 certifies, with the frozen lexicon instrument applied to the whole corpus, that the outraged statement carries anger- and fear-laden vocabulary and the measured statement carries none (Table 16).

## 6.3 Sample Waterfall

Table 8 traces the Experiment 2 sample from the raw survey export to the three registered exclusion regimes, following the study's archived exclusion log. Entrants who failed the pre-treatment audio check twice were screened out before random assignment and therefore do not appear among the randomized completes. Within the randomized pool, 3,838 respondents (99.8 percent) had passed the audio check on the first attempt and 7 on the second, and the difference between the primary regime and the first-attempt regime corresponds to those seven respondents. The export contained no platform spam-flagged completions, and

the registration's deviations log records that this platform flag is not among the registered exclusions.

Table 8: Sample waterfall for Experiment 2 (dynamic design), from raw survey export to the three registered exclusion regimes

| | N |
|---|---|
| *From export to the randomized pool* | |
| Rows in the raw survey export | 3,866 |
| Preview/test rows removed | 2 |
| Completed responses | 3,864 |
| Exited before random assignment (declined consent or screened out) | 18 |
| Duplicate Prolific ID (first submission retained) | 1 |
| Completed in under 90 seconds | 0 |
| Randomized completed responses | 3,845 |
| *Checks within the randomized pool* | |
| Passed the audio check on the first attempt | 3,838 |
| Passed the audio check on the second attempt | 7 |
| Passed the instructed-response attention check | 3,601 |
| *Registered exclusion regimes* | |
| Primary (attention passers) | 3,601 |
| No attention exclusion | 3,845 |
| First-attempt audio passers | 3,594 |

*Note*: Attention passers are 93.7 percent of the 3,845 randomized completed responses and first-attempt audio passers 99.8 percent; the three registered regimes retain 93.7, 100.0, and 93.5 percent of that pool, respectively. The 18 unrandomized exits reached the end of the survey without an assigned condition, having declined consent or been screened out, and are removed under the registered plan. Source: exclusions.txt in the study's analysis archive.

## 6.4 Full Results

Hypothesis labels follow the pre-analysis plan. Its H2a and H2b are the manuscript's third and fourth hypotheses, and its H4 is the manuscript's second. Tables 9 and 10 report every estimand of Experiment 2's registered analysis for perceived health expertise and external efficacy under the primary exclusion regime, for both specifications and all three respondent groups. Table 11 reports the same estimands for the three mechanism batteries in the full sample. Table 12 reports the confirmatory quantities under all three registered exclusion regimes, and Table 13 the registered equivalence (TOST) details for the two bounded quantities, under both the registered fixed bound and the analyzed-sample-SD sensitivity bound. Full estimand tables under the two alternative regimes are available in the registered analysis archives. The manipulation checks reported in the main text are estimated in the same framework. The outraged register raised the seven-point emotional-intensity check by $\beta = 1.850$ (95% CI $[1.765, 1.935]$, $p < .001$), and respondents overwhelmingly identified their assigned format correctly: 96.5 percent of those assigned a solo condition reported the MC appeared alone, and 93.3 percent of those assigned a split-screen reported the MC appeared alongside another person. Table 15 reports per-cell means and standard deviations for every outcome under every registered exclusion regime.

Table 9: Experiment 2, full coefficient estimates, primary registered exclusion regime: perceived health expertise (benchmarked). The solo-minus-split display companions (`solo_advantage_*`) are exact sign reversals of the split-minus-solo rows shown and are omitted.

| | ATE | | | | CTE | | | |
|---|---|---|---|---|---|---|---|---|
| Subgroup | $n$ | $\beta$ | 95% CI | $p$ | $n$ | $\beta$ | 95% CI | $p$ |
| *Woman MC: split-screen − solo, pooled over registers (H1, registered one-sided)* (`h1_female_visual`) | | | | | | | | |
| All | 3,601 | 0.009 | [−0.062, 0.079] | .813 (.594) | 3,594 | −0.000 | [−0.071, 0.071] | .997 (.498) |
| Women | 1,786 | 0.051 | [−0.047, 0.149] | .306 (.847) | 1,784 | 0.043 | [−0.054, 0.141] | .385 (.807) |
| Men | 1,765 | −0.045 | [−0.150, 0.059] | .393 (.196) | 1,761 | −0.053 | [−0.157, 0.051] | .320 (.160) |
| *Man MC: split-screen − solo, pooled over registers* (`h1_male_companion`) | | | | | | | | |
| All | 3,601 | −0.045 | [−0.130, 0.040] | .300 | 3,594 | −0.041 | [−0.126, 0.043] | .337 |
| Women | 1,786 | −0.072 | [−0.192, 0.048] | .240 | 1,784 | −0.055 | [−0.174, 0.065] | .367 |
| Men | 1,765 | −0.036 | [−0.160, 0.088] | .571 | 1,761 | −0.037 | [−0.160, 0.086] | .558 |
| *Visual × register interaction, woman MC (H2a, registered one-sided)* (`h2a_visXlang_female`) | | | | | | | | |
| All | 3,601 | −0.030 | [−0.172, 0.111] | .675 (.663) | 3,594 | −0.016 | [−0.157, 0.125] | .825 (.587) |
| Women | 1,786 | 0.010 | [−0.186, 0.206] | .918 (.459) | 1,784 | 0.041 | [−0.154, 0.235] | .682 (.341) |
| Men | 1,765 | −0.083 | [−0.292, 0.126] | .436 (.782) | 1,761 | −0.072 | [−0.280, 0.136] | .497 (.752) |
| *Gender × register interaction (H2b)* (`h2b_genderXlang`) | | | | | | | | |
| All | 3,601 | −0.067 | [−0.178, 0.044] | .236 | 3,594 | −0.069 | [−0.179, 0.041] | .219 |
| Women | 1,786 | 0.015 | [−0.140, 0.169] | .854 | 1,784 | 0.005 | [−0.149, 0.159] | .952 |
| Men | 1,765 | −0.173 | [−0.335, −0.011] | .036 | 1,761 | −0.165 | [−0.327, −0.004] | .045 |
| *Three-way interaction (exploratory)* (`threeway_interaction`) | | | | | | | | |
| All | 3,601 | −0.157 | [−0.379, 0.065] | .165 | 3,594 | −0.149 | [−0.370, 0.072] | .186 |
| Women | 1,786 | −0.157 | [−0.466, 0.153] | .321 | 1,784 | −0.125 | [−0.432, 0.182] | .425 |
| Men | 1,765 | −0.208 | [−0.532, 0.116] | .209 | 1,761 | −0.208 | [−0.531, 0.115] | .207 |
| *Outraged − measured register, both MCs* (`outrage_effect_all`) | | | | | | | | |
| All | 3,601 | −0.025 | [−0.081, 0.030] | .374 | 3,594 | −0.025 | [−0.080, 0.030] | .371 |
| Women | 1,786 | −0.032 | [−0.109, 0.045] | .416 | 1,784 | −0.031 | [−0.108, 0.046] | .426 |
| Men | 1,765 | −0.018 | [−0.099, 0.063] | .657 | 1,761 | −0.012 | [−0.093, 0.069] | .767 |
| *Outraged − measured register, woman MC* (`outrage_effect_female`) | | | | | | | | |
| All | 3,601 | −0.059 | [−0.130, 0.012] | .105 | 3,594 | −0.060 | [−0.131, 0.011] | .098 |
| Women | 1,786 | −0.025 | [−0.123, 0.073] | .619 | 1,784 | −0.029 | [−0.126, 0.069] | .562 |
| Men | 1,765 | −0.105 | [−0.209, −0.001] | .048 | 1,761 | −0.095 | [−0.199, 0.009] | .075 |
| *Outraged − measured register, man MC* (`outrage_effect_male`) | | | | | | | | |
| All | 3,601 | 0.008 | [−0.077, 0.094] | .847 | 3,594 | 0.009 | [−0.075, 0.094] | .829 |
| Women | 1,786 | −0.039 | [−0.159, 0.080] | .519 | 1,784 | −0.034 | [−0.152, 0.085] | .580 |
| Men | 1,765 | 0.068 | [−0.056, 0.192] | .280 | 1,761 | 0.070 | [−0.053, 0.194] | .265 |
| *Outraged − measured, woman MC solo (H3 diagnostic)* (`h3_outrage_solo_female`) | | | | | | | | |
| All | 3,601 | −0.074 | [−0.174, 0.027] | .150 | 3,594 | −0.068 | [−0.168, 0.032] | .184 |
| Women | 1,786 | −0.020 | [−0.163, 0.123] | .787 | 1,784 | −0.009 | [−0.151, 0.133] | .906 |
| Men | 1,765 | −0.147 | [−0.288, −0.005] | .043 | 1,761 | −0.131 | [−0.272, 0.010] | .069 |
| *Outraged − measured, woman MC split-screen* (`h3_outrage_split_female`) | | | | | | | | |
| All | 3,601 | −0.043 | [−0.144, 0.057] | .396 | 3,594 | −0.052 | [−0.152, 0.048] | .310 |
| Women | 1,786 | −0.030 | [−0.164, 0.104] | .661 | 1,784 | −0.049 | [−0.182, 0.084] | .470 |
| Men | 1,765 | −0.064 | [−0.217, 0.089] | .416 | 1,761 | −0.059 | [−0.212, 0.095] | .454 |
| *Woman MC: split-screen − solo, measured register (exploratory)* (`split_female_measured`) | | | | | | | | |
| All | 3,601 | −0.007 | [−0.109, 0.096] | .900 | 3,594 | −0.008 | [−0.111, 0.095] | .877 |
| Women | 1,786 | 0.056 | [−0.084, 0.196] | .432 | 1,784 | 0.064 | [−0.076, 0.203] | .373 |
| Men | 1,765 | −0.087 | [−0.240, 0.066] | .266 | 1,761 | −0.089 | [−0.243, 0.065] | .257 |
| *Woman MC: split-screen − solo, outraged register (exploratory)* (`split_female_outraged`) | | | | | | | | |
| All | 3,601 | 0.024 | [−0.074, 0.122] | .634 | 3,594 | 0.008 | [−0.089, 0.105] | .874 |
| Women | 1,786 | 0.046 | [−0.091, 0.183] | .510 | 1,784 | 0.023 | [−0.113, 0.159] | .740 |
| Men | 1,765 | −0.004 | [−0.146, 0.138] | .956 | 1,761 | −0.017 | [−0.157, 0.124] | .815 |

## 6.5 Exploratory Causal-Mediation Estimates

Our pre-analysis plan designated causal-mediation analysis as exploratory, because its estimates require the assumption of sequential ignorability, an assumption that randomizing the treatments does not guarantee for the mediators. Table 14 reports average causal mediation effects (ACME), average direct effects (ADE), and total effects for the three paths discussed in the main text, estimated on the woman-MC subsample under the primary regime ($n = 2{,}161$) with quasi-Bayesian confidence intervals (300 simulations) and robust standard errors. The table also reports the sensitivity analysis these estimates require, namely the value of the

Table 10: Experiment 2, full coefficient estimates, primary registered exclusion regime: external efficacy. The split-minus-solo companions (`h1_*`) are exact sign reversals of the solo-advantage rows shown and are omitted.

| | ATE | | | | CTE | | | |
|---|---|---|---|---|---|---|---|---|
| Subgroup | $n$ | $\beta$ | 95% CI | $p$ | $n$ | $\beta$ | 95% CI | $p$ |
| *Man MC: solo − split-screen (display orientation)* (`solo_advantage_male`) | | | | | | | | |
| All | 3,601 | 0.016 | [−0.127, 0.159] | .823 | 3,594 | 0.036 | [−0.101, 0.173] | .608 |
| Women | 1,786 | −0.113 | [−0.316, 0.089] | .273 | 1,784 | −0.022 | [−0.216, 0.172] | .826 |
| Men | 1,765 | 0.117 | [−0.088, 0.323] | .263 | 1,761 | 0.082 | [−0.113, 0.278] | .410 |
| *Woman MC: solo − split-screen (display orientation; H4 on external efficacy)* (`h4_solo_advantage_female`) | | | | | | | | |
| All | 3,601 | 0.070 | [−0.046, 0.187] | .237 | 3,594 | 0.060 | [−0.052, 0.173] | .294 |
| Women | 1,786 | 0.158 | [−0.005, 0.321] | .057 | 1,784 | 0.136 | [−0.021, 0.293] | .090 |
| Men | 1,765 | −0.016 | [−0.186, 0.153] | .849 | 1,761 | −0.009 | [−0.173, 0.155] | .915 |
| *Difference in solo advantages: woman MC − man MC* (`solo_advantage_diff`) | | | | | | | | |
| All | 3,601 | 0.054 | [−0.131, 0.238] | .567 | 3,594 | 0.024 | [−0.153, 0.202] | .787 |
| Women | 1,786 | 0.271 | [0.011, 0.531] | .041 | 1,784 | 0.158 | [−0.092, 0.408] | .216 |
| Men | 1,765 | −0.134 | [−0.400, 0.132] | .324 | 1,761 | −0.091 | [−0.346, 0.164] | .485 |
| *Visual × register interaction, woman MC (H2a, registered one-sided)* (`h2a_visXlang_female`) | | | | | | | | |
| All | 3,601 | −0.130 | [−0.363, 0.103] | .274 | 3,594 | −0.170 | [−0.396, 0.055] | .138 |
| Women | 1,786 | −0.144 | [−0.470, 0.182] | .386 | 1,784 | −0.229 | [−0.542, 0.085] | .153 |
| Men | 1,765 | −0.103 | [−0.442, 0.235] | .550 | 1,761 | −0.112 | [−0.440, 0.217] | .505 |
| *Gender × register interaction (H2b)* (`h2b_genderXlang`) | | | | | | | | |
| All | 3,601 | −0.057 | [−0.241, 0.128] | .547 | 3,594 | −0.061 | [−0.238, 0.117] | .503 |
| Women | 1,786 | −0.158 | [−0.418, 0.102] | .234 | 1,784 | −0.172 | [−0.421, 0.076] | .174 |
| Men | 1,765 | 0.015 | [−0.252, 0.281] | .914 | 1,761 | 0.040 | [−0.214, 0.295] | .755 |
| *Three-way interaction (exploratory)* (`threeway_interaction`) | | | | | | | | |
| All | 3,601 | −0.103 | [−0.472, 0.266] | .584 | 3,594 | −0.132 | [−0.486, 0.223] | .466 |
| Women | 1,786 | 0.086 | [−0.434, 0.606] | .745 | 1,784 | 0.004 | [−0.494, 0.503] | .987 |
| Men | 1,765 | −0.274 | [−0.807, 0.258] | .312 | 1,761 | −0.260 | [−0.771, 0.252] | .321 |
| *Outraged − measured register, both MCs* (`outrage_effect_all`) | | | | | | | | |
| All | 3,601 | 0.050 | [−0.042, 0.143] | .284 | 3,594 | 0.048 | [−0.041, 0.136] | .292 |
| Women | 1,786 | 0.018 | [−0.112, 0.148] | .789 | 1,784 | 0.003 | [−0.121, 0.128] | .959 |
| Men | 1,765 | 0.083 | [−0.050, 0.217] | .219 | 1,761 | 0.095 | [−0.032, 0.223] | .143 |
| *Outraged − measured register, woman MC* (`outrage_effect_female`) | | | | | | | | |
| All | 3,601 | 0.022 | [−0.094, 0.138] | .711 | 3,594 | 0.017 | [−0.095, 0.130] | .763 |
| Women | 1,786 | −0.061 | [−0.224, 0.102] | .463 | 1,784 | −0.083 | [−0.240, 0.074] | .301 |
| Men | 1,765 | 0.091 | [−0.078, 0.260] | .293 | 1,761 | 0.116 | [−0.049, 0.280] | .168 |
| *Outraged − measured register, man MC* (`outrage_effect_male`) | | | | | | | | |
| All | 3,601 | 0.079 | [−0.064, 0.222] | .281 | 3,594 | 0.078 | [−0.059, 0.215] | .264 |
| Women | 1,786 | 0.097 | [−0.106, 0.299] | .350 | 1,784 | 0.089 | [−0.103, 0.282] | .364 |
| Men | 1,765 | 0.076 | [−0.129, 0.282] | .468 | 1,761 | 0.075 | [−0.120, 0.270] | .450 |
| *Outraged − measured, woman MC solo (H3 diagnostic)* (`h3_outrage_solo_female`) | | | | | | | | |
| All | 3,601 | −0.043 | [−0.207, 0.121] | .606 | 3,594 | −0.068 | [−0.226, 0.090] | .399 |
| Women | 1,786 | −0.133 | [−0.367, 0.100] | .264 | 1,784 | −0.197 | [−0.423, 0.029] | .087 |
| Men | 1,765 | 0.039 | [−0.192, 0.270] | .740 | 1,761 | 0.060 | [−0.162, 0.281] | .596 |
| *Outraged − measured, woman MC split-screen* (`h3_outrage_split_female`) | | | | | | | | |
| All | 3,601 | 0.087 | [−0.079, 0.253] | .303 | 3,594 | 0.103 | [−0.058, 0.263] | .211 |
| Women | 1,786 | 0.011 | [−0.217, 0.238] | .925 | 1,784 | 0.031 | [−0.186, 0.249] | .778 |
| Men | 1,765 | 0.142 | [−0.105, 0.390] | .259 | 1,761 | 0.171 | [−0.071, 0.414] | .166 |
| *Woman MC: split-screen − solo, measured register (exploratory)* (`split_female_measured`) | | | | | | | | |
| All | 3,601 | −0.135 | [−0.302, 0.031] | .112 | 3,594 | −0.146 | [−0.307, 0.016] | .078 |
| Women | 1,786 | −0.230 | [−0.463, 0.002] | .052 | 1,784 | −0.250 | [−0.474, −0.026] | .028 |
| Men | 1,765 | −0.035 | [−0.278, 0.207] | .776 | 1,761 | −0.047 | [−0.284, 0.190] | .698 |
| *Woman MC: split-screen − solo, outraged register (exploratory)* (`split_female_outraged`) | | | | | | | | |
| All | 3,601 | −0.005 | [−0.168, 0.157] | .950 | 3,594 | 0.025 | [−0.132, 0.181] | .755 |
| Women | 1,786 | −0.086 | [−0.315, 0.142] | .460 | 1,784 | −0.022 | [−0.242, 0.198] | .846 |
| Men | 1,765 | 0.068 | [−0.168, 0.304] | .572 | 1,761 | 0.065 | [−0.163, 0.292] | .577 |

sensitivity parameter $\rho$ at which each ACME would equal zero under violations of sequential ignorability. The contestation battery is the least reliable of our measures ($\alpha = 0.44$), so the first path should be read with particular caution.

# 7 The Emotional Register of the Corpus

The register manipulation in Experiment 2 compares an outraged and a measured version of the same position. The main text motivates that manipulation in part observationally,

Table 11: Experiment 2, full coefficient estimates for the three mechanism batteries, full sample, primary registered exclusion regime (ATE, with CTE in the row below each estimand)

| | Perceived warmth | | | Norm violation | | | Contestation | | |
|---|---|---|---|---|---|---|---|---|---|
| Estimand | $\beta$ | 95% CI | $p$ | $\beta$ | 95% CI | $p$ | $\beta$ | 95% CI | $p$ |
| `h1_female_visual` | 0.012 | [−0.095, 0.119] | .827 | −0.090 | [−0.196, 0.015] | .092 | −0.085 | [−0.163, −0.007] | .033 |
| CTE | −0.001 | [−0.103, 0.101] | .988 | −0.082 | [−0.181, 0.018] | .107 | −0.086 | [−0.163, −0.009] | .028 |
| `h1_male_companion` | −0.069 | [−0.214, 0.076] | .351 | 0.006 | [−0.098, 0.111] | .903 | −0.048 | [−0.142, 0.046] | .320 |
| CTE | −0.054 | [−0.196, 0.088] | .455 | −0.005 | [−0.105, 0.095] | .922 | −0.049 | [−0.142, 0.044] | .303 |
| `solo_advantage_male` | 0.069 | [−0.076, 0.214] | .351 | −0.006 | [−0.111, 0.098] | .903 | 0.048 | [−0.046, 0.142] | .320 |
| CTE | 0.054 | [−0.088, 0.196] | .455 | 0.005 | [−0.095, 0.105] | .922 | 0.049 | [−0.044, 0.142] | .303 |
| `h4_solo_advantage_female` | −0.012 | [−0.119, 0.095] | .827 | 0.090 | [−0.015, 0.196] | .092 | 0.085 | [0.007, 0.163] | .033 |
| CTE | 0.001 | [−0.101, 0.103] | .988 | 0.082 | [−0.018, 0.181] | .107 | 0.086 | [0.009, 0.163] | .028 |
| `solo_advantage_diff` | −0.081 | [−0.261, 0.099] | .378 | 0.097 | [−0.051, 0.245] | .200 | 0.037 | [−0.085, 0.159] | .554 |
| CTE | −0.053 | [−0.228, 0.122] | .550 | 0.077 | [−0.065, 0.218] | .288 | 0.037 | [−0.084, 0.158] | .551 |
| `h2a_visXlang_female` | −0.001 | [−0.215, 0.213] | .991 | 0.032 | [−0.179, 0.242] | .767 | 0.001 | [−0.155, 0.157] | .987 |
| CTE | −0.035 | [−0.239, 0.169] | .738 | 0.063 | [−0.136, 0.261] | .534 | 0.010 | [−0.144, 0.163] | .903 |
| `h2b_genderXlang` | −0.268 | [−0.448, −0.087] | .004 | 0.335 | [0.187, 0.483] | < .001 | 0.031 | [−0.092, 0.153] | .624 |
| CTE | −0.265 | [−0.440, −0.090] | .003 | 0.330 | [0.189, 0.471] | < .001 | 0.031 | [−0.090, 0.151] | .619 |
| `threeway_interaction` | −0.049 | [−0.409, 0.311] | .790 | 0.120 | [−0.176, 0.416] | .427 | −0.004 | [−0.249, 0.240] | .972 |
| CTE | −0.033 | [−0.382, 0.316] | .854 | 0.119 | [−0.163, 0.401] | .408 | 0.011 | [−0.231, 0.252] | .932 |
| `outrage_effect_all` | −0.281 | [−0.371, −0.191] | < .001 | 1.027 | [0.953, 1.101] | < .001 | 0.434 | [0.373, 0.495] | < .001 |
| CTE | −0.286 | [−0.373, −0.198] | < .001 | 1.035 | [0.965, 1.106] | < .001 | 0.439 | [0.378, 0.499] | < .001 |
| `outrage_effect_female` | −0.415 | [−0.522, −0.308] | < .001 | 1.194 | [1.089, 1.299] | < .001 | 0.449 | [0.371, 0.527] | < .001 |
| CTE | −0.418 | [−0.520, −0.316] | < .001 | 1.200 | [1.101, 1.300] | < .001 | 0.454 | [0.377, 0.531] | < .001 |
| `outrage_effect_male` | −0.147 | [−0.292, −0.003] | .046 | 0.859 | [0.755, 0.963] | < .001 | 0.419 | [0.324, 0.513] | < .001 |
| CTE | −0.153 | [−0.295, −0.012] | .034 | 0.870 | [0.770, 0.970] | < .001 | 0.423 | [0.330, 0.517] | < .001 |
| `h3_outrage_solo_female` | −0.416 | [−0.568, −0.263] | < .001 | 1.210 | [1.063, 1.358] | < .001 | 0.450 | [0.342, 0.558] | < .001 |
| CTE | −0.435 | [−0.582, −0.289] | < .001 | 1.232 | [1.092, 1.372] | < .001 | 0.459 | [0.353, 0.565] | < .001 |
| `h3_outrage_split_female` | −0.414 | [−0.565, −0.264] | < .001 | 1.178 | [1.028, 1.329] | < .001 | 0.449 | [0.336, 0.561] | < .001 |
| CTE | −0.401 | [−0.544, −0.257] | < .001 | 1.169 | [1.029, 1.309] | < .001 | 0.449 | [0.339, 0.560] | < .001 |
| `split_female_measured` | 0.011 | [−0.135, 0.158] | .880 | −0.074 | [−0.193, 0.044] | .219 | −0.084 | [−0.193, 0.025] | .132 |
| CTE | −0.018 | [−0.160, 0.123] | .800 | −0.050 | [−0.164, 0.063] | .387 | −0.081 | [−0.190, 0.028] | .145 |
| `split_female_outraged` | 0.013 | [−0.143, 0.168] | .874 | −0.106 | [−0.280, 0.068] | .231 | −0.085 | [−0.197, 0.026] | .133 |
| CTE | 0.017 | [−0.131, 0.164] | .825 | −0.113 | [−0.276, 0.050] | .173 | −0.091 | [−0.199, 0.017] | .100 |

Table 12: Experiment 2, confirmatory quantities under all three registered exclusion regimes (ATE)

| Quantity | Regime | $n$ | $\beta$ | 95% CI | $p$ |
|---|---|---|---|---|---|
| H1: woman MC visual contrast | Primary (attention rule) | 3,601 | 0.009 | [−0.062, 0.079] | .813 (.594) |
| | No attention exclusion | 3,845 | 0.005 | [−0.064, 0.074] | .881 (.560) |
| | First-attempt audio only | 3,594 | 0.009 | [−0.063, 0.080] | .814 (.593) |
| H2a: visual × register, woman MC | Primary (attention rule) | 3,601 | −0.030 | [−0.172, 0.111] | .675 (.663) |
| | No attention exclusion | 3,845 | −0.030 | [−0.168, 0.109] | .674 (.663) |
| | First-attempt audio only | 3,594 | −0.028 | [−0.170, 0.114] | .699 (.651) |
| H2b: gender × register (competence) | Primary (attention rule) | 3,601 | −0.067 | [−0.178, 0.044] | .236 |
| | No attention exclusion | 3,845 | −0.097 | [−0.205, 0.011] | .080 |
| | First-attempt audio only | 3,594 | −0.064 | [−0.175, 0.047] | .257 |
| H2b: gender × register (warmth) | Primary (attention rule) | 3,601 | −0.268 | [−0.448, −0.087] | .004 |
| | No attention exclusion | 3,845 | −0.251 | [−0.425, −0.078] | .005 |
| | First-attempt audio only | 3,594 | −0.264 | [−0.445, −0.084] | .004 |
| H3 diagnostic: outrage in woman MC's solo | Primary (attention rule) | 3,601 | −0.074 | [−0.174, 0.027] | .150 |
| | No attention exclusion | 3,845 | −0.085 | [−0.183, 0.013] | .089 |
| | First-attempt audio only | 3,594 | −0.070 | [−0.171, 0.030] | .170 |
| H4: woman MC solo advantage, efficacy, women | Primary (attention rule) | 1,786 | 0.158 | [−0.005, 0.321] | .057 |
| | No attention exclusion | 1,904 | 0.139 | [−0.017, 0.295] | .081 |
| | First-attempt audio only | 1,780 | 0.160 | [−0.003, 0.323] | .055 |

Table 13: Experiment 2, registered equivalence (TOST) details for the two bounded quantities, all regimes and both bound sources

| Quantity | Regime | Bound source | $\beta$ | $\Delta$ | $p_{\text{lower}}$ | $p_{\text{upper}}$ | Verdict |
|---|---|---|---|---|---|---|---|
| H3 diagnostic (outrage, solo; competence) | Primary | registered | −0.074 | 0.110 | .243 | < .001 | not established |
| | Primary | analyzed SD | −0.074 | 0.126 | .155 | < .001 | not established |
| H4 (solo advantage, efficacy, women) | Primary | registered | 0.158 | 0.199 | < .001 | .311 | not established |
| | Primary | analyzed SD | 0.158 | 0.205 | < .001 | .286 | not established |
| H3 diagnostic (outrage, solo; competence) | No attention excl. | registered | −0.085 | 0.110 | .311 | < .001 | not established |
| | No attention excl. | analyzed SD | −0.085 | 0.127 | .201 | < .001 | not established |
| H4 (solo advantage, efficacy, women) | No attention excl. | registered | 0.139 | 0.199 | < .001 | .226 | not established |
| | No attention excl. | analyzed SD | 0.139 | 0.204 | < .001 | .210 | not established |
| H3 diagnostic (outrage, solo; competence) | First-attempt audio | registered | −0.070 | 0.110 | .223 | < .001 | not established |
| | First-attempt audio | analyzed SD | −0.070 | 0.126 | .139 | < .001 | not established |
| H4 (solo advantage, efficacy, women) | First-attempt audio | registered | 0.160 | 0.199 | < .001 | .319 | not established |
| | First-attempt audio | analyzed SD | 0.160 | 0.205 | < .001 | .293 | not established |

Table 14: Exploratory causal-mediation estimates for Experiment 2 (woman-MC subsample, primary exclusion regime), with sensitivity analyses

| Quantity | Estimate | 95% CI | $p$ |
|---|---|---|---|
| *Path 1: split-screen → perceived contestation → health expertise* | | | |
| ACME | 0.010 | [0.001, 0.020] | .040 |
| ADE | −0.003 | [−0.071, 0.066] | .913 |
| Total effect | 0.006 | [−0.061, 0.075] | .833 |
| $\rho$ at which ACME = 0 | −0.2 | | |
| *Path 2a: outraged register → perceived warmth → health expertise* | | | |
| ACME | −0.038 | [−0.053, −0.025] | < .001 |
| ADE | −0.021 | [−0.100, 0.049] | .500 |
| Total effect | −0.059 | [−0.141, 0.014] | .107 |
| $\rho$ at which ACME = 0 | 0.2 | | |
| *Path 2b: outraged register → perceived norm violation → health expertise* | | | |
| ACME | −0.082 | [−0.117, −0.047] | < .001 |
| ADE | 0.023 | [−0.061, 0.099] | .593 |
| Total effect | −0.059 | [−0.129, 0.006] | .080 |
| $\rho$ at which ACME = 0 | −0.2 | | |

*Note*: Quasi-Bayesian estimates (300 simulations) with robust standard errors, woman-MC subsample under the primary regime (n = 2,161); ACME is the average causal mediation effect, ADE the average direct effect. Sensitivity rows report the error-correlation $\rho$ at which the ACME equals zero; the corresponding $\tilde{R}^2_M \tilde{R}^2_Y$ products are 0.039, 0.038, and 0.032. Source: the study's archived mediation output.

reporting that the measured register is the most common presentation in our corpus. This section documents that claim and describes the instrument behind it. By 'outrage' we mean the register described in studies of political opinion media: presentation designed to provoke a visceral response, dramatizing threat and grievance rather than weighing positions (Sobieraj and Berry, 2011). The outraged script of Experiment 2 treats this register as a combination of anger and fear ("outraged, terrified, and forgotten"), so the appropriate lexical instrument is one that codes the two emotions separately.

We use the NRC Emotion Lexicon (EmoLex, version 0.92), a crowdsourced word–emotion association lexicon (Mohammad and Turney, 2013), taking its anger category (1,245 single-word English headwords in the version used here) and its fear category (1,474 headwords).

Table 15: Experiment 2, per-cell means and standard deviations, full analyzed sample, every outcome and registered exclusion regime

| Regime | Condition | $n$ | Health exp. | | Ext. efficacy | | Warmth | | Norm viol. | | Contest. | |
|---|---|---|---|---|---|---|---|---|---|---|---|---|
| | | | m | SD | m | SD | m | SD | m | SD | m | SD |
| Primary regime | Woman MC, solo, measured | 542 | 0.744 | 0.877 | −0.419 | 1.391 | 1.263 | 1.255 | −1.763 | 0.980 | −0.632 | 0.918 |
| | Woman MC, solo, outraged | 537 | 0.670 | 0.804 | −0.462 | 1.351 | 0.847 | 1.298 | −0.553 | 1.444 | −0.182 | 0.887 |
| | Woman MC, split, measured | 536 | 0.738 | 0.844 | −0.554 | 1.401 | 1.274 | 1.202 | −1.838 | 1.007 | −0.716 | 0.913 |
| | Woman MC, split, outraged | 546 | 0.694 | 0.838 | −0.467 | 1.378 | 0.860 | 1.319 | −0.659 | 1.475 | −0.268 | 0.982 |
| | Man MC, solo, measured | 355 | 0.525 | 0.836 | −0.562 | 1.431 | 0.837 | 1.370 | −1.709 | 0.889 | −0.543 | 0.853 |
| | Man MC, solo, outraged | 352 | 0.597 | 0.829 | −0.497 | 1.394 | 0.713 | 1.475 | −0.894 | 1.131 | −0.121 | 0.920 |
| | Man MC, split, measured | 367 | 0.543 | 0.824 | −0.592 | 1.355 | 0.792 | 1.284 | −1.747 | 0.816 | −0.588 | 0.875 |
| | Man MC, split, outraged | 366 | 0.488 | 0.809 | −0.500 | 1.357 | 0.620 | 1.466 | −0.843 | 1.158 | −0.172 | 0.992 |
| No attention excl. | Woman MC, solo, measured | 577 | 0.730 | 0.877 | −0.415 | 1.384 | 1.256 | 1.248 | −1.723 | 1.004 | −0.598 | 0.926 |
| | Woman MC, solo, outraged | 579 | 0.645 | 0.820 | −0.410 | 1.329 | 0.841 | 1.283 | −0.513 | 1.422 | −0.164 | 0.887 |
| | Woman MC, split, measured | 574 | 0.721 | 0.845 | −0.516 | 1.406 | 1.284 | 1.191 | −1.772 | 1.070 | −0.691 | 0.917 |
| | Woman MC, split, outraged | 578 | 0.666 | 0.846 | −0.442 | 1.377 | 0.865 | 1.316 | −0.655 | 1.464 | −0.245 | 0.983 |
| | Man MC, solo, measured | 380 | 0.499 | 0.845 | −0.531 | 1.433 | 0.851 | 1.369 | −1.661 | 0.934 | −0.518 | 0.857 |
| | Man MC, solo, outraged | 385 | 0.575 | 0.821 | −0.489 | 1.390 | 0.687 | 1.479 | −0.864 | 1.127 | −0.125 | 0.916 |
| | Man MC, split, measured | 388 | 0.526 | 0.836 | −0.574 | 1.350 | 0.796 | 1.275 | −1.705 | 0.841 | −0.549 | 0.878 |
| | Man MC, split, outraged | 384 | 0.503 | 0.824 | −0.486 | 1.347 | 0.630 | 1.462 | −0.820 | 1.158 | −0.163 | 0.979 |
| First-attempt audio | Woman MC, solo, measured | 540 | 0.742 | 0.877 | −0.419 | 1.393 | 1.262 | 1.257 | −1.764 | 0.981 | −0.635 | 0.918 |
| | Woman MC, solo, outraged | 536 | 0.672 | 0.804 | −0.458 | 1.350 | 0.849 | 1.298 | −0.552 | 1.445 | −0.182 | 0.887 |
| | Woman MC, split, measured | 535 | 0.736 | 0.844 | −0.555 | 1.402 | 1.273 | 1.202 | −1.838 | 1.008 | −0.717 | 0.914 |
| | Woman MC, split, outraged | 545 | 0.694 | 0.839 | −0.465 | 1.379 | 0.861 | 1.320 | −0.660 | 1.477 | −0.271 | 0.980 |
| | Man MC, solo, measured | 355 | 0.525 | 0.836 | −0.562 | 1.431 | 0.837 | 1.370 | −1.709 | 0.889 | −0.543 | 0.853 |
| | Man MC, solo, outraged | 351 | 0.595 | 0.829 | −0.502 | 1.392 | 0.709 | 1.476 | −0.892 | 1.132 | −0.119 | 0.920 |
| | Man MC, split, measured | 366 | 0.543 | 0.825 | −0.592 | 1.357 | 0.790 | 1.286 | −1.748 | 0.817 | −0.587 | 0.876 |
| | Man MC, split, outraged | 366 | 0.488 | 0.809 | −0.500 | 1.357 | 0.620 | 1.466 | −0.843 | 1.158 | −0.172 | 0.992 |

These two categories were chosen because they separate our own scripts. Under the matching rule stated below, the outraged script contains anger- and fear-laden vocabulary and the measured script contains none, so the instrument that certifies the manipulation can be carried to the corpus unchanged. As a robustness check we also apply the moral-emotional dictionary of Brady et al. (2017), comprising 18 whole-word entries and 54 prefix stems, tokens of which appear in neither script.

The matching rule is fixed as follows. Caption text and scripts alike are lowercased, apostrophes are deleted in place, and tokens are maximal runs of the letters a through z, with all other characters treated as separators. A token matches an EmoLex category if it equals a headword exactly or if, ending in -ed, -d, or -ing with a residual base of at least three characters, one of the candidates formed by the bare base, the base with a restored final e, or the base with a doubled final consonant reduced equals a headword. No plural stripping is performed, no other suffixes are handled, and no prefix or infix matching is attempted. Tokens are matched against the moral-emotional dictionary without any stemming, a token counting if it equals a whole-word entry or begins with one of the stems. The rule strips only verbal and participial suffixes. Exact matching alone would fail to count the outraged script's "outraged" (the headword is "outrage"), while stripping plurals as well would count the measured script's neutral "challenges." The tradeoff is that some plurals are missed ("threats" does not match "threat"), which biases every corpus count below downward. The rule also makes no attempt to handle negation, so "not anxious" would count toward fear.

Table 16 reports every lexicon hit in the two scripts, scoring only the MC's spoken statement in each register, since the anchor's introduction and closing are common to all eight conditions. The outraged script (36 tokens) contains one anger match ("outraged," via the headword "outrage") and two fear matches ("anxious" and "forgotten," both exact), and the measured script (39 tokens) contains no match of any kind in any of the three lexicons, so

the separation the manipulation requires (anger- and fear-laden vocabulary on one side, none on the other) holds with no incidental matches to explain away. It is important to note that "terrified" matches nothing, because version 0.92 contains neither "terrified" nor "terrify". In this word family the fear coding is assigned to "terror" and "terrorize," and no reasonable one-word rule connects "terrified" to those forms. Because the gap is in the lexicon and not in the scripts, the corpus counts below, if anything, understate the register they measure.

Table 16: Lexicon matches in the two Experiment-2 scripts under the frozen matching rule

| Script | Tokens | Anger matches | Fear matches | Moral-emotional matches |
|---|---|---|---|---|
| Outraged | 36 | "outraged" | "anxious," "forgotten" | none |
| Measured | 39 | none | none | none |

*Note*: Only the MC's spoken statement is scored in each register; the anchor's introduction and closing are common to all eight conditions and are not scored. Token counts are under the tokenization stated in the text. "Outraged" matches the anger headword "outrage" under the inflection rule; "anxious" and "forgotten" are exact fear headwords; "terrified" matches no entry in any of the three lexicons (see text). Anger and fear are the EmoLex categories of Mohammad and Turney (2013); moral-emotional is the dictionary of Brady et al. (2017). Source: the frozen-rule script scoring archived with the study's analysis materials.

Table 17 applies the same instrument to the full corpus. At least one anger- or fear-connoting token appears in 92.5 percent of the 374,643 CNN segments, 93.7 percent of the 320,821 Fox News segments, and 93.1 percent of the 695,464 segments combined. This is not surprising, since a typical minute of captions runs roughly 137 words and is scored against more than 2,000 distinct headwords. For this reason we focus on the rate. Anger- and fear-laden vocabulary amounts to 28.3 tokens per 1,000 words in the average CNN segment and 29.8 in the average Fox News segment (combined mean 29.0, median 25.6), so roughly 97 percent of the words in a typical minute of this coverage carry neither code. The moral-emotional dictionary shows the same pattern, with at least one token in 62.8 percent of segments combined. In other words, emotional vocabulary is present in the typical minute of this coverage, but it does not define the overall register. The measured presentation is the most common one on both networks, and the outraged register is comparatively rare.

The same conclusion follows from the individual words the two scripts use to mark their registers. Across the combined corpus, "outraged" appears in 0.27 percent of segments, "terrified" in 0.20 percent, "anxious" in 0.10 percent, and "forgotten" in 0.36 percent,[1] while the measured script's marker words are likewise rare ("focused," 0.73 percent; "committed," 0.80 percent; "solutions," 0.33 percent; "overlooked," 0.04 percent; "unaccounted," 0.04 percent; "unaddressed," 0.002 percent).

These counts have three limits. First, each segment mixes anchors, guests, correspondents, and voice-over without speaker identification, so they characterize the discourse environment and attribute nothing to any individual speaker. Second, the count files carry no shot-format labels of their own, so the solo-versus-split comparison below rests entirely on

[1] "Terrified" and "forgotten" entered the outraged script in the first place because the word-frequency analysis of Section 8.3 found them among the frequent words when congresswomen discuss women on these networks. Their appearance both there and in the script is therefore by construction, so their prevalence here illustrates the pattern but does not independently validate the script.

Table 17: Anger- and fear-laden vocabulary in the full corpus, by network

| | CNN | Fox News | Combined |
|---|---|---|---|
| Segments (N) | 374,643 | 320,821 | 695,464 |
| ≥1 anger-or-fear token (% of segments) | 92.5 | 93.7 | 93.1 |
| ≥1 anger token (%) | 82.2 | 84.6 | 83.3 |
| ≥1 fear token (%) | 88.2 | 89.6 | 88.9 |
| Anger-or-fear tokens per 1,000 words (mean) | 28.3 | 29.8 | 29.0 |
| Anger-or-fear tokens per 1,000 words (median) | 25.0 | 26.5 | 25.6 |
| ≥1 moral-emotional token (%) | 61.0 | 64.9 | 62.8 |

*Note*: Each cell reports the share of segments whose paired minute of closed-caption text contains at least one token of the named category under the matching rule stated in the text; Ns are the corpus totals of the Data and Methods section. Numerators for the anger-or-fear row are 346,488 (CNN), 300,675 (Fox News), and 647,163 (combined). The rate rows are per-segment rates of anger-or-fear tokens per 1,000 words, averaged (or taken at the median) across segments; five CNN segments whose caption files contain no alphabetic tokens are excluded from the rate rows only, where a rate is undefined, and count as zeros in every percentage row. The moral-emotional row counts the Brady et al. (2017) dictionary. Source: the per-segment register counts archived with the study's analysis materials (`register_counts_cnn.csv.gz`, `register_counts_fox.csv.gz`).

the join to the shot classifications. Third, the matching rule is a fixed lexical instrument with the negation and plural-recall limits stated above.

## 7.1 The Register of the Coverage by Shot Type

This subsection joins the same per-segment counts to the shot classifications underlying Figure 1 of the main text, so that the register of the coverage can be compared across visual formats and the outraged script judged against the distributions those formats produce. The five categories are constructed exactly as in that figure's replication code. A frame is a solo appearance when the facial-recognition pipeline identified exactly one legislator in it (solo woman or solo man), and a frame in which two legislators were identified enters one of three split-screen categories (a woman and a man, two women, or two men) when the CLIP pipeline of the Data and Methods section classified it as a split-screen. These are the 226,067 solo and 53,177 split-screen frames of the main text. Two-legislator frames the pipeline did not classify as split-screens (4,984 on CNN and 8,762 on Fox News) and frames with three or more identified legislators (14,021 and 15,323) belong to no category, exactly as in the figure. (The replication code applies the split-screen filter only when assigning the two-legislator categories; the asymmetry is moot, as no solo-classified frame appears in either network's split-screen list.)

Frames join to caption segments by their shared broadcast-and-minute stem. No cell contains a duplicate filename, so classified frames map one-to-one onto the per-segment counts of the corpus analysis above, and the join is essentially complete. Every CNN cell joins at 100 percent, and no Fox News cell joins below 99.93 percent. The 81 unmatched Fox News frames trace to a handful of special broadcasts (an inauguration and several one-off specials) whose caption files are absent from the corpus text, and every joined segment contains at least one alphabetic token. Table 18 reports the joined cells.

Two patterns are relevant to the register manipulation. First, the register varies only

Table 18: Anger- and fear-laden vocabulary by shot type

| Shot type | Segments | ≥1 anger-or-fear token (%) | Rate per 1,000 words Mean | Median | Marker words (% of segments) outraged | terrified | anxious | forgotten |
|---|---|---|---|---|---|---|---|---|
| *CNN* | | | | | | | | |
| Solo woman | 30,390 | 94.4 | 28.3 | 25.1 | 0.20 | 0.22 | 0.16 | 0.28 |
| Solo man | 70,426 | 94.5 | 28.3 | 25.0 | 0.20 | 0.16 | 0.12 | 0.30 |
| Split: woman and man | 10,688 | 95.4 | 26.4 | 23.7 | 0.36 | 0.20 | 0.14 | 0.28 |
| Split: two women | 2,458 | 94.3 | 26.4 | 23.7 | 0.33 | 0.12 | 0.12 | 0.37 |
| Split: two men | 14,512 | 95.7 | 26.7 | 24.0 | 0.22 | 0.25 | 0.12 | 0.20 |
| *Fox News* | | | | | | | | |
| Solo woman | 43,336 | 95.1 | 29.3 | 26.1 | 0.46 | 0.20 | 0.09 | 0.51 |
| Solo man | 81,842 | 95.2 | 29.5 | 26.3 | 0.34 | 0.18 | 0.09 | 0.64 |
| Split: woman and man | 10,968 | 96.0 | 28.2 | 25.4 | 0.35 | 0.20 | 0.08 | 0.52 |
| Split: two women | 3,304 | 95.9 | 28.4 | 25.3 | 0.58 | 0.24 | 0.03 | 0.70 |
| Split: two men | 11,239 | 96.0 | 28.8 | 25.8 | 0.44 | 0.14 | 0.12 | 0.71 |

*Note*: Each row is a shot-type cell of Figure 1 of the main text joined to the per-segment register counts behind Table 17. Segments is the number of classified frames whose caption segment joined: the five CNN cells join at 100 percent (30,390; 70,426; 10,688; 2,458; and 14,512 frames), and the five Fox News cells join at 99.93 percent or better (43,366; 81,885; 10,970; 3,304; and 11,245 frames, of which 30, 43, 2, 0, and 6 lack caption files; see text). Rate columns are per-segment rates of anger-or-fear tokens per 1,000 words under the matching rule stated in the text, averaged (or taken at the median) across segments; no joined segment lacks alphabetic tokens. Marker-word columns report the share of segments containing the named word at least once, counted exactly as the corpus-wide prevalences reported in the text ("terrified" as the literal token, independent of its absence from the lexicon). Source: the by-shot merge archived with the study's analysis materials (`register_by_shot.csv`, `register_by_shot_segments.csv.gz`).

modestly across shot types. Anger- and fear-laden vocabulary is somewhat denser in the minutes surrounding solo appearances than in those surrounding split-screens. Pooling the networks, the rate is 28.9 words per 1,000 when a congresswoman appears alone against 27.3 when she shares the screen with a congressman and 27.5 when two congresswomen share it. These differences are reliable (the 95 percent confidence intervals do not overlap) but never larger than two words per 1,000, and Fox News is roughly one to two words per 1,000 above CNN in every shot type. Second, the coverage shows essentially no gender difference in this register. Pooling the networks, the mean rate in a congresswoman's solo minutes differs from a congressman's by −0.06 words per 1,000 (95% CI $[-0.23, 0.11]$, $p = .456$; within CNN −0.02, $p = .865$; within Fox News −0.19, $p = .094$), and the share of solo minutes containing at least one of the four words with which the outraged script marks its register ("outraged," "terrified," "anxious," "forgotten") is 1.10 percent for congresswomen and 1.03 percent for congressmen (difference 0.07 percentage points, $p = .125$; null within each network, $p = .141$ and $p = .808$). This suggests the discourse environment surrounding congresswomen's appearances is not more emotionally charged than the one surrounding congressmen's, whatever individual legislators say on screen.

We also use this join to ask how common the outraged script's register is in the corpus. Under the frozen matching rule the script contains 3 matched words in 36 tokens, a rate of 83.3 per 1,000 applied here as the exact fraction 3/36. Across the 695,459 corpus segments with at least one token, 12,317 (1.77 percent) match or exceed that rate, so the script's concentration of anger- and fear-laden vocabulary exceeds that of 98.2 percent of corpus segments (98.3 percent on CNN, 98.1 percent on Fox News), and it is rarer still in the solo cells, where 1.37 percent of a congresswoman's solo minutes reach the script's rate and 1.46

percent of a congressman's. We also classified each segment by the nearer script anchor, using the midpoint of 41.7 words per 1,000 (halfway between the two scripts' densities of 0 and 83.3). By this classification, 539,447 of the 695,459 tokened segments (77.6 percent) fall on the measured side. Taken together, this suggests the outraged script is an extreme but realistic example of this coverage. Its rate is roughly three times the on-air mean, matched or exceeded by only one to two segments in a hundred, and no more characteristic of the minutes surrounding congresswomen's appearances than of congressmen's.

These joined counts have the same limits stated above, plus one more. The five categories cover the legislator-identified subset of the corpus, 279,163 of the 695,464 segments, so their denominators differ from those of Table 17, and the frames outside the five categories are excluded by construction. And because each segment's minute of captions mixes anchors, guests, correspondents, and voice-over, the cell differences describe the discourse environment surrounding each shot type, not what the legislators on screen, including congresswomen, actually said.

# 8 The Pilot Studies

## 8.1 The 2024 Pilot Waves

The fully crossed experiment reported as Experiment 1 in the main text extends a simpler two-condition design that we fielded twice in 2024, first on a national Qualtrics panel and then on a student sample. Both waves presented only the woman MC, appearing either alone or beside her man colleague, so they could not establish whether the penalty they document is gendered. The fully crossed design of Experiment 1 addresses that question. Both waves are fully reported here. Table 2 provides the demographic breakdown of these samples alongside the December 2024 pilot's samples and the two 2026 experimental samples.

The first wave was a pre-registered online vignette experiment administered through Qualtrics. Respondents (N=925, 54% women) completed the survey in February 2024.[2] After giving informed consent and answering demographic, partisanship, and voting-history questions, respondents were randomly assigned to one of two treatment conditions: selected frames from a CNN cable news program showing a woman MC either alone or with her man colleague in a mixed-gender dyad, with the chyron "Congress Set to Support Frontline Healthcare Providers" and a brief healthcare vignette. Responses to the healthcare item were standardized against the mean evaluation of three other policy areas (education, national security, and the economy).

We report both average (ATE) and conditional treatment effects (CTE). In both instances, a simple linear regression was used to estimate the effect of treatment. In the latter, we included controls for whether the participant had a college degree, identified as a Democrat, or identified as White. These variables were included to adjust for demographic differences across our survey samples (see Table 2). With this in mind, for the Qualtrics sample we find the ATE associated with our treatment is positive and statistically significant ($\beta = 0.177, t = 2.928, df = 942, p < 0.004$), implying congresswomen are perceived as being

[2]The pre-revision manuscript's prose reported N = 944 (55% women) and a February 15–19 field period for this wave, while its Table 1 and both figure notes reported N = 925 (54% women) fielded February 1–14. We report the machine-generated figures here. The regression statistics quoted below for this wave (df = 942 and 937) are as originally reported and reflect the pre-revision run.

more competent when they appear by themselves, as opposed to when they appear in a split-screen. The result holds when the effect of treatment is conditioned on demographic controls ($\beta = 0.175, t = 2.876, df = 937, p < 0.005$). The same can be said for both the ATE and CTE when the data are subset to women participants (N = 500). However, no significant effect was found when only men were included when estimating the treatment effect (N = 425).

To demonstrate the robustness of this finding, we replicated this result using a small student sample (N = 195). The replication was conducted April 12–26, 2024, at a large Midwestern research university, with students receiving extra credit to participate. In the student sample, the ATE associated with our treatment is still positive and statistically significant ($\beta = 0.467, t = 3.073, df = 193, p < 0.003$), as well as the CTE ($\beta = 0.476, t = 3.107, df = 190, p < 0.003$). Again, with the student sample we found the coefficient associated with treatment for both the ATE and CTE was significant when the sample was subset to women (N = 107), whereas the same cannot be said for men (N = 88). Finally, when pooled together with the Qualtrics sample, the coefficient associated with treatment is statistically significant in both the ATE ($\beta = 0.119, t = 2.764, df = 1118, p < 0.006$) and CTE ($\beta = 0.115, t = 2.653, df = 1114, p < 0.009$), providing strong support for our first hypothesis and suggesting split-screen appearances tend to significantly decrease the perceived expertise of congresswomen.

Although these results strongly support the first hypothesis, the external-efficacy estimates provide little evidence for the second. We expected that women viewers exposed to solo shots of congresswomen would report higher levels of external efficacy, reflecting the commanding presence such visuals convey. External efficacy was measured as the mean of three items, with higher values indicating greater efficacy. Across all model specifications, the treatment coefficient was never statistically significant at the 0.05 level, though its direction was generally consistent with expectations.

All four pilot samples saw the pre-revision static frames (a solo woman MC, or the woman MC beside her man colleague; the December 2024 pilot superimposed closed captioning on the same frames, Figure 4), read a healthcare vignette, and completed the same capability and efficacy items as the 2026 experiments (Section 5.4), both batteries administered on five-point scales in these pilots (the 2026 experiments used a seven-point response scale for the efficacy items), with the assigned frame re-displayed during the outcome questions with a green box around the MC being evaluated. The December 2024 registration, which both 2026 pre-analysis plans extend, is credited in Section 8.2.

As reported in Table 19, no significant imbalance was found between treatment and control for any variable except `College Degree` and `Democrat`, and for those two only in the Qualtrics sample (and, given its relative size, in the combined sample). Both variables are controlled in all CTE results reported in this section, so this imbalance should not overly influence the estimates.

## 8.2 The December 2024 Pilot and Its MTurk Replication

Between the 2024 pilot waves reported in Section 8.1 and the fully crossed design of Experiment 1, we fielded a further pre-registered study asking whether the split-screen penalty persists when the message is emotionally charged. In the revised manuscript this study is reported here in full. It is relevant to the main text in three ways. First, it replicates the

Table 19: Balance tests for the 2024 pilot waves

| | Qualtrics (q) | | | | | Students (s) | | | | | Both (b) | | | | |
|---|---|---|---|---|---|---|---|---|---|---|---|---|---|---|---|
| Variable | $\bar{T}_q$ | $\bar{C}_q$ | $t_q$ | $df_q$ | $p_q$ | $\bar{T}_s$ | $\bar{C}_s$ | $t_s$ | $df_s$ | $p_s$ | $\bar{T}_b$ | $\bar{C}_b$ | $t_b$ | $df_b$ | $p_b$ |
| College Degree | 0.45 | 0.52 | -2.28 | 923 | 0.02 | 0.11 | 0.09 | 0.36 | 193 | 0.72 | 0.39 | 0.45 | -1.96 | 1118 | 0.05 |
| Democrat | 0.43 | 0.37 | 2.11 | 923 | 0.04 | 0.47 | 0.45 | 0.33 | 193 | 0.74 | 0.44 | 0.38 | 2.04 | 1118 | 0.04 |
| White | 0.59 | 0.63 | -1.17 | 923 | 0.24 | 0.66 | 0.56 | 1.48 | 193 | 0.14 | 0.60 | 0.61 | -0.44 | 1118 | 0.66 |
| Women | 0.56 | 0.51 | 1.57 | 923 | 0.12 | 0.53 | 0.58 | -0.75 | 193 | 0.45 | 0.55 | 0.52 | 1.11 | 1118 | 0.27 |
| Healthcare | 0.68 | 0.66 | 0.41 | 923 | 0.68 | 1.04 | 1.05 | -0.07 | 193 | 0.95 | 0.74 | 0.73 | 0.33 | 1118 | 0.74 |
| Security | 0.15 | 0.24 | -1.36 | 923 | 0.17 | 0.45 | 0.75 | -2.40 | 193 | 0.02 | 0.20 | 0.33 | -2.16 | 1118 | 0.03 |
| Education | 0.52 | 0.56 | -0.66 | 923 | 0.51 | 0.77 | 0.92 | -1.29 | 193 | 0.20 | 0.57 | 0.63 | -1.11 | 1118 | 0.27 |
| Economy | 0.33 | 0.42 | -1.39 | 923 | 0.16 | 0.60 | 0.81 | -1.71 | 193 | 0.09 | 0.38 | 0.49 | -1.94 | 1118 | 0.05 |

*Note*: This table reports the results from a series of t-tests comparing the means of variables under treatment and control. Variable names are provided in the first column. Treatment ($\bar{T}$) and control ($\bar{C}$) means are then reported in the next two columns, followed by the $t$ statistic ($t$) and the associated degrees of freedom ($df$) and p-value ($p$). Results based on our Qualtrics and Student samples include $q$ and $s$ in the subscripts, respectively. Combined results are presented with the $b$ subscript. All $t$ tests are two-sided and equal variance is assumed.

Figure 4: Example of a CNN broadcast image used in the December 2024 study, where a woman MC appears alongside her man colleague and closed captioning is superimposed

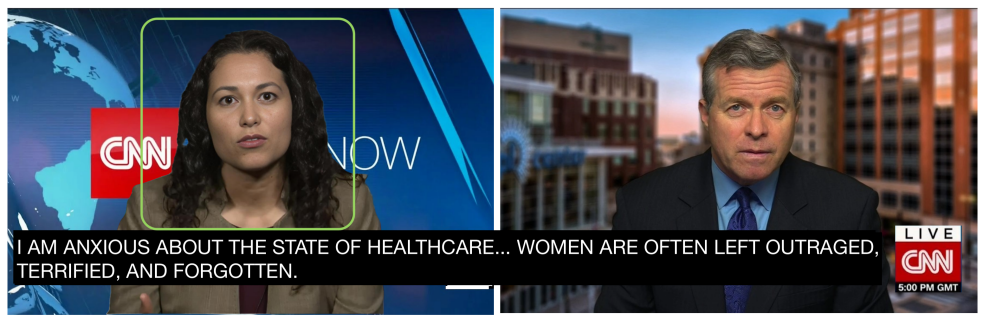


split-screen penalty on two additional platforms. Second, it shows that the penalty survives the presence of substantive, emotionally charged on-screen text, which rules out on-screen text as an explanation for the null result of Experiment 2. Third, because its emotionally charged register appeared in both conditions, it motivates the randomized register manipulation of Experiment 2.

This study extends the 2024 waves' tests by considering the emotional register surrounding women's issues. Prior work shows that when the topic is women, congresswomen speak with heightened emotional intensity (Dietrich et al., 2019) and that women legislators are more likely to use emotive language (Gennaro and Ash, 2022). Parallel research on digital appeals documents increased use of outrage among women candidates (Russell et al., 2024), even as classic experiments warn of backlash to women's anger (Okimoto and Brescoll, 2010). Consistent with this, the word-frequency analysis of Section 8.3 shows that when congresswomen discuss women on CNN and Fox News, the most characteristic completions include outrage-related terms (e.g., "terrified," "forgotten").

We do not claim that split-screen versus solo shots cause congresswomen to change their rhetoric, because guests are generally unaware of how they are featured on the screen. Rather, expressions of outrage may co-occur with both solo and split-screen shots, ultimately shaping competence judgments. Because women who express anger or moral outrage are evaluated more negatively than comparable men (Marshburn et al., 2020; Okimoto and Brescoll, 2010; Brescoll, 2016; Hargrave, 2023), we expected lower perceived competence when viewers encountered congresswomen expressing outrage.

To test whether the split-screen penalty persists when the message is emotionally charged,

we pre-registered a further online vignette experiment administered through Prolific.[3] Respondents (N=500, 52% women) completed the survey on December 10, 2024, and after consent and the same demographic battery as the earlier waves were randomly assigned to one of two treatment conditions. These treatments used the same two frames as the waves reported in Section 8.1: the woman MC appearing alone, or beside her male counterpart in a mixed-gender dyad. As shown in Figure 4, in both conditions we also superimposed closed captioning which read: "I AM ANXIOUS ABOUT THE STATE OF HEALTHCARE. WOMEN ARE OFTEN LEFT OUTRAGED, TERRIFIED, AND FORGOTTEN." We chose this sentence because it incorporated the words most characteristic of congresswomen's descriptions of women when they appeared by themselves (see Section 8.3). Because the caption was identical across the two conditions, the emotionally charged register was held constant, meaning the design can reveal whether the visual penalty survives the presence of outrage on the screen, though not what outrage itself contributes.

We report both average (ATE) and conditional treatment effects (CTE), specified exactly as in the waves above. For the Prolific sample we find the ATE associated with our treatment is positive and statistically significant ($\beta = 0.142, t = 1.996, df = 498, p < 0.047$), implying congresswomen are perceived as being more competent when they appear by themselves, as opposed to when they appear in a split-screen. The result holds when the effect of treatment is conditioned on the demographic controls ($\beta = 0.142, t = 2.007, df = 49, p < 0.046$)[4]. However, no significant effect was found when the data was subset to women ($ATE\colon \beta = 0.161, t = 1.675, df = 258, p < 0.096; CTE\colon \beta = 0.174, t = 1.840, df = 255, p < 0.067$) and men ($ATE\colon \beta = 0.143, t = 1.350, df = 230, p < 0.179; CTE\colon \beta = 0.149, t = 1.393, df = 227, p < 0.165$) respondents.

To demonstrate the robustness of this finding, we replicated this result using a small sample obtained from Amazon's Mechanical Turk (N = 130). In the MTurk replication, the ATE associated with our treatment is still positive and statistically significant ($\beta = 0.336, t = 2.040, df = 128, p < 0.044$), as well as the CTE ($\beta = 0.332, t = 2.000, df = 126, p < 0.048$). Again, we find the coefficient associated with treatment was no longer statistically significant when the data was subset to women (N = 34) and men (N = 96), even though the coefficient was higher in the former as opposed to the latter. Finally, when pooled together with the Prolific sample, the coefficient associated with treatment is statistically significant in both the ATE ($\beta = 0.186, t = 2.786, df = 628, p < 0.006$) and CTE ($\beta = 0.182, t = 2.783, df = 624, p < 0.006$), providing further support for our first hypothesis. When the pooled data was subset to women ($ATE\colon \beta = 0.182, t = 1.954, df = 292, p < 0.052; CTE\colon \beta = 0.196, t = 2.134, df = 288, p < 0.034$) and men ($ATE\colon \beta = 0.199, t = 2.079, df = 326, p < 0.039; CTE\colon \beta = 0.184, t = 1.968, df = 322, p < 0.050$) significant or marginally significant effects were found for the coefficient associated with

[3] This study was registered with the Open Science Framework in December 2024. The pre-analysis plans for the two experiments reported in the main text extend that registration, and their anonymized view-only links appear in the corresponding main-text footnotes. The December 2024 registration itself has been public since June 2025 and therefore cannot be linked anonymously. A direct link will be provided in the published version.

[4] The degrees of freedom printed here carry a typographical error from the original manuscript (approximately 494 expected given the sample and controls). The coefficient, $t$ statistic, and $p$ value are as originally reported.

treatment. Unlike the waves reported in Section 8.1, where the penalty was concentrated among women respondents, the pooled effect here does not differ appreciably by respondent gender.

The pre-revision manuscript also observed that the coefficient associated with treatment in this study ranged from 0.14 to 0.34, whereas the same coefficient ranged from 0.10 to 0.21 in the 2024 waves,[5] and read this comparison as suggesting the negatively valenced closed-captioning amplified the visual penalty. Because the caption was identical in both conditions, this was a comparison between designs rather than randomized groups and could not support a causal claim. The randomized register manipulation in Experiment 2 of the main text finds no such amplification of the competence penalty, so the difference in coefficient ranges is best attributed to differences in samples and platforms, not to the on-screen text.

We again found little support for our second hypothesis. Across all model specifications, the external-efficacy treatment coefficient was never statistically significant at the 0.05 level, though its direction was generally consistent with expectations.

Table 20: Balance tests for the December 2024 pilot and its MTurk replication

| | Prolific (r) | | | | | MTurk (m) | | | | | Both (b) | | | | |
|---|---|---|---|---|---|---|---|---|---|---|---|---|---|---|---|
| Variable | $\bar{\bar{T}}_r$ | $\overline{C}_r$ | $t_r$ | $df_r$ | $p_r$ | $\bar{\bar{T}}_m$ | $\overline{C}_m$ | $t_m$ | $df_m$ | $p_m$ | $\bar{\bar{T}}_b$ | $\overline{C}_b$ | $t_b$ | $df_b$ | $p_b$ |
| College Degree | 0.67 | 0.63 | 0.75 | 532 | 0.45 | 1.00 | 0.97 | 1.42 | 142 | 0.16 | 0.73 | 0.70 | 0.74 | 657 | 0.46 |
| Democrat | 0.36 | 0.39 | -0.56 | 532 | 0.58 | 0.49 | 0.54 | -0.66 | 142 | 0.51 | 0.39 | 0.42 | -0.79 | 657 | 0.43 |
| White | 0.66 | 0.65 | 0.30 | 532 | 0.76 | 0.93 | 0.89 | 0.87 | 142 | 0.39 | 0.72 | 0.70 | 0.56 | 657 | 0.58 |
| Women | 0.52 | 0.51 | 0.17 | 532 | 0.86 | 0.29 | 0.24 | 0.75 | 142 | 0.45 | 0.48 | 0.45 | 0.81 | 657 | 0.42 |
| Healthcare | 0.77 | 0.71 | 0.71 | 501 | 0.48 | 0.61 | 0.36 | 1.31 | 142 | 0.19 | 0.75 | 0.63 | 1.46 | 628 | 0.14 |
| Security | 0.12 | 0.22 | -1.22 | 500 | 0.22 | 0.35 | 0.40 | -0.30 | 142 | 0.76 | 0.17 | 0.25 | -1.09 | 627 | 0.28 |
| Education | 0.39 | 0.43 | -0.52 | 501 | 0.60 | 0.38 | 0.35 | 0.15 | 142 | 0.88 | 0.39 | 0.41 | -0.21 | 628 | 0.83 |
| Economy | 0.24 | 0.34 | -1.14 | 500 | 0.25 | 0.32 | 0.43 | -0.64 | 142 | 0.52 | 0.26 | 0.36 | -1.27 | 627 | 0.20 |

*Note*: Columns as in Table 19, with the Prolific and MTurk samples denoted by the $r$ and $m$ subscripts, respectively, and combined results by the $b$ subscript. All $t$ tests are two-sided and equal variance is assumed.

As reported in Table 20, no significant imbalance was found for any variable, including `College Degree` and `Democrat`, suggesting the Qualtrics sample was an outlier in that regard. Both variables are again controlled in all CTE results reported in this section.

## 8.3 Word Frequencies When Congresswomen Appear Alone

The closed captioning superimposed on the December 2024 pilot's stimuli was drawn from a masked-language-model analysis of the corpus. Using BERT (Devlin et al., 2019), we fine-tuned a model on the closed-caption segments in which our pipeline identified a single congresswoman on screen and asked it to complete "Women are [MASK]," benchmarking each word's probability against the foundational Google News model (reported as the percent increase over that baseline). Table 21 reports the five most likely completions on each network. Negative emotion words such as *terrified* and *forgotten* top both lists, and the December 2024 pilot's caption incorporates them verbatim.

[5] As stated in the pre-revision manuscript, the student replication's coefficients (0.467 ATE, 0.476 CTE) fall outside this range.

Table 21: Congresswomen are more likely to say women are "terrified" and "forgotten" when they appear by themselves on CNN and Fox News

| **CNN** | %Δ | **FOX** | %Δ |
|---|---|---|---|
| terrified | 4598.87 | forgotten | 194.32 |
| forgotten | 789.55 | terrified | 158.98 |
| horrible | 742.84 | horrible | 106.09 |
| appalled | 594.87 | americans | 91.57 |
| awful | 396.64 | insulted | 89.69 |

*Note*: This table reports the results from a MLM model estimated using the CNN and Fox News data collected for this study. Results are reported as a percent increase over the foundational model, which was trained on Google News data. All percents are positive, suggesting the five words listed in each column are *more* likely to appear in our data than the Google News data. Each word is used to fill in the blank in the following sentence: "Women are ___________."



\end{document}